\documentclass[pdflatex,sn-mathphys-num]{sn-jnl}% Math and Physical Sciences Numbered Reference Style
\usepackage{graphicx}%
\usepackage{multirow}%
\usepackage{amsmath,amssymb,amsfonts}%
\usepackage{amsthm}%
\usepackage{mathrsfs}%
\usepackage[title]{appendix}%
\usepackage{xcolor}%
\usepackage{textcomp}%
\usepackage{manyfoot}%
\usepackage{booktabs}%
\usepackage{algorithm}%
\usepackage{algorithmicx}%
\usepackage{algpseudocode}%
\usepackage{listings}%
\usepackage{natbib}
\usepackage{tcolorbox}
\usepackage{devanagari}
\usepackage{rotating}
\usepackage{makecell}
\theoremstyle{thmstyleone}%
\theoremstyle{thmstyletwo}%

\theoremstyle{thmstylethree}%

\begin{document}

\title[Bridging the Linguistic Divide: A Survey on Leveraging Large Language Models for Machine Translation]{Bridging the Linguistic Divide: A Survey on Leveraging Large Language Models for Machine Translation}


\author[1]{\fnm{Baban} \sur{Gain}}\email{gainbaban@gmail.com}

\author[1]{\fnm{Dibyanayan} \sur{Bandyopadhayay}}\email{dibyanayan@gmail.com}

\author[1]{\fnm{Asif} \sur{Ekbal}}\email{asif@iitp.ac.in}

\author[2]{\fnm{Trilok Nath} \sur{Singh}}\email{tnsiitb@gmail.com}

\affil[1]{\orgdiv{Department of Computer Science and Engineering}, \orgname{IIT Patna}, \orgaddress{\postcode{801106}, \country{India}}}

\affil[2]{\orgdiv{Department of Civil and Environmental Engineering}, \orgname{IIT Patna}, \orgaddress{\postcode{801106}, \country{India}}}


\abstract{Large Language Models (LLMs) are rapidly reshaping machine translation (MT), particularly by introducing instruction-following, in-context learning, and preference-based alignment into what has traditionally been a supervised encoder–decoder paradigm. This survey provides a comprehensive and up-to-date overview of how LLMs are being leveraged for MT across data regimes, languages, and application settings. We systematically analyze prompting-based methods, parameter-efficient and full fine-tuning strategies, synthetic data generation, preference-based optimization, and reinforcement learning with human and weakly supervised feedback. Special attention is given to low-resource translation, where we examine the roles of synthetic data quality, diversity, and preference signals, as well as the limitations of current RLHF pipelines. We further review recent advances in Mixture-of-Experts models, MT-focused LLMs, and multilingual alignment, highlighting trade-offs between scalability, specialization, and accessibility. Beyond sentence-level translation, we survey emerging document-level and discourse-aware MT methods with LLMs, showing that most approaches extend sentence-level pipelines through structured context selection, post-editing, or reranking rather than requiring fundamentally new data regimes or architectures. Finally, we discuss LLM-based evaluation, its strengths and biases, and its role alongside learned metrics. Overall, this survey positions LLM-based MT as an evolution of traditional MT systems, where gains increasingly depend on data quality, preference alignment, and context utilization rather than scale alone, and outlines open challenges for building robust, inclusive, and controllable translation systems.}




\maketitle

\section{Introduction}

Machine Translation (MT) has significantly evolved from early rule-based systems to advanced neural models. Initially, MT relied heavily on manually crafted linguistic rules and dictionaries, requiring extensive human intervention for each language pair and struggling with linguistic nuances, idiomatic expressions, and contextual understanding \cite{HUTCHINS1995431}. The transition to Statistical MT (SMT) in the late 1980s and early 1990s marked a major advancement, as SMT utilized probabilistic models trained on large bilingual corpora, automating translation processes \cite{brown-etal-1993-mathematics, Koehn_2009}. However, SMT faced challenges such as limited fluency, difficulties with less common language pairs, and issues handling long-range dependencies.
The introduction of neural networks marked a turning point in MT research. Recurrent Neural Networks (RNNs) and later, sequence-to-sequence models with Long Short-Term Memory (LSTM) \cite{10.1162/neco.1997.9.8.1735} units improved translation fluency by learning contextual dependencies \cite{cho-etal-2014-learning,10.5555/2969033.2969173}. However, these architectures faced limitations in handling long sequences due to vanishing gradients and computational constraints. The emergence of attention mechanisms \cite{bahdanau2016neuralmachinetranslationjointly,luong-etal-2015-effective} addressed these issues by allowing models to focus selectively on relevant parts of the input sequence, enhancing both accuracy and coherence. This led to the development of the Transformer model \cite{NIPS2017_3f5ee243}, which became the foundation for modern neural MT systems. MT is now widely integrated into global communication from business collaborations to cross-cultural teamwork enabling multilingual interactions in real time or asynchronously. A study \cite{10.1145/3512937} show that MT-mediated exchanges significantly enhance communication clarity, discussion depth, and decision-making quality in multilingual team settings. 

Despite advancements in MT, translating low-resource languages remains a significant challenge. Most high-performing MT models rely on extensive parallel corpora, which are scarce for many languages. Researchers have explored various strategies, including transfer learning \cite{zoph-etal-2016-transfer,gu-etal-2018-universal,kim-etal-2019-effective,Ji_Zhang_Duan_Zhang_Chen_Luo_2020,aji-etal-2020-neural}, multilingual training \cite{firat-etal-2016-zero},  data augmentation \cite{fadaee-etal-2017-data,xia-etal-2019-generalized}, etc., to improve translation quality for low-resource languages. These approaches have shown promise but still face difficulties in capturing linguistic diversity, domain adaptation, and ensuring consistent translation quality.

Challenges in MT extend beyond data scarcity. Linguistic ambiguity, polysemy \cite{ohuoba-etal-2024-quantifying}, domain-specific terminology, and cultural nuances \cite{yao-etal-2024-benchmarking,NAVEEN2024110878} continue to pose obstacles to accurate translation. Additionally, hallucination, where the model generates plausible yet incorrect  \cite{dale-etal-2023-detecting,10.1162/tacl_a_00615}, and robustness against adversarial attacks \cite{zhang-etal-2021-crafting,sadrizadeh2023transfooladversarialattackneural,10095342} remain active areas of research.

LLMs offer new opportunities to address these challenges. Their ability to learn from vast amounts of multilingual and monolingual text allows them to generate translations with improved fluency and contextual awareness. Few-shot and zero-shot learning capabilities enable LLMs to handle low-resource languages with minimal supervision. Moreover, LLMs facilitate domain adaptation and personalized translation by leveraging in-context learning. However, they also introduce concerns such as increased computational costs, potential biases, and the necessity for efficient deployment strategies.

As MT continues to evolve, the integration of LLMs is poised to fundamentally transform the field, introducing new capabilities while also raising important ethical and computational considerations. Although prior surveys have reviewed various aspects of MT \cite{10438431,10112530,haddow-etal-2022-survey}, to the best of our knowledge, none have systematically captured the substantial advancements enabled by LLMs in recent years. Given the diverse range of techniques available ranging from prompting strategies to parameter-efficient fine-tuning, navigating this emerging landscape can be challenging for newcomers and practitioners alike. To address this gap, this paper presents a comprehensive analysis of the principal methodologies for leveraging LLMs in MT, offering detailed comparisons and insights to guide future research and application.

\section{Structure of the Survey}
We followed a structured but lightweight PRISMA-style process to identify and organize relevant literature on LLMs for MT. Our search was conducted using Google Scholar, ACL Anthology, arXiv, and Google Search. The search queries included, but were not limited to, keywords such as ``machine translation with LLMs'', ``prompting for MT'', ``LLM-based evaluation'', ``fine-tuning LLMs for translation'', ``synthetic data generation for MT'', ``low-resource translation with LLMs'', and ``LLM vs encoder-decoder MT''. For section-specific coverage, we additionally used targeted keywords such as ``in-context learning for MT'', ``LoRA/QLoRA for translation'', ``hallucination in MT'', ``LLM-as-a-judge'', ``agentic workflows for translation'', and ``document-level translation with LLMs''. When using Google Search, we explicitly appended the term ``research paper'' to the query to filter out non-academic content such as blogs, tutorials, or product pages.

We primarily targeted papers published from 2022 onwards, corresponding to the period in which LLM-based MT research gained significant momentum. To expand coverage beyond the initial keyword-based search, we examined the abstracts of retrieved papers to identify recurring terminologies and methodological patterns, which were then used as additional search cues to discover related work. We further employed backward and forward citation tracking starting from influential papers. In cases where multiple studies reported highly similar methodologies or convergent conclusions, we prioritized inclusion based on the earliest publication venue or arXiv submission date, and cited later works selectively when they provided substantive extensions or additional insights. We exclude the papers which is specific for a narrow language or domain, unless strong justifications are present about its generalizability.

Initially  we organized the literature into four high-level categories: purely prompting-based approaches, fine-tuning-based approaches, LLM-based evaluation methods, and comparative studies between LLMs and encoder-decoder MT models with some papers covering multiple aspects. To better capture methodological overlap and thematic diversity, we further introduced sub-groupings within each category, such as zero-shot and few-shot prompting, parameter-efficient fine-tuning, preference-based optimization, and agentic or multi-step workflows. While certain studies employ hybrid techniques such as reinforcement learning (RL) combined with quantized low-rank adaptation (QLoRA) they are categorized under the primary methodology most representative of their core contribution.  The paper is structured as follows:

\begin{itemize}
    \item \textbf{Section III: Synthetic Data Generation with LLMs} – This section explores how LLMs can be leveraged to generate synthetic bilingual data, a critical technique for improving MT performance in data-scarce scenarios. It reviews various strategies, including back-translation, forward translation, and domain adaptation techniques.
    
    \item \textbf{Section IV: Translation Methodologies} – Here, we categorize MT approaches using LLMs into two primary paradigms: (1) \textit{Prompting-Based Techniques}, which leverage pre-trained LLMs without modifying their parameters, and (2) \textit{Fine-Tuning Approaches}, which adapt LLMs to specific translation tasks by modifying their weights. We provide an in-depth analysis of methods such as zero-shot prompting, in-context learning, supervised fine-tuning, and parameter-efficient fine-tuning techniques like LoRA.
    
    \item \textbf{Section V: LLMs vs Encoder-Decoder across Languages} – This section evaluates how LLMs perform across different languages, with a special focus on low-resource language pairs.
    
    \item \textbf{Section VI: LLM based Evaluation} – The final technical section discusses various evaluation metrics and methodologies used to assess LLM-driven MT. 
    \item \textbf{Section VII: MT Beyond Sentence-level} – Here we provide a brief study of methods that use additional context to translate the sentences including document translation and literary translation. 

\end{itemize}

This study offers a comprehensive analysis of how LLMs are being used for MT. It highlights the emerging challenges and opportunities, providing a foundation for future research in this rapidly evolving domain. To ensure methodological consistency, the study refrains from direct cross-paper comparisons, as existing works vary significantly in terms of datasets, language pairs, and evaluation metrics.

For clarity and focus, the analysis is restricted to text-to-text, sentence-level translation, while including only a short discussion on document-level\cite{10.1145/3441691,wang-etal-2023-document-level} while specialized domains such as dialogue-based, and multimodal approaches \cite{shen2024surveymultimodalmachinetranslation}, sign language translation \cite{NUNEZMARCOS2023118993}) has been excluded. 

\section{Synthetic Data Generation with LLMs}
\label{sec:data-gen}
Domain adaptation in MT has been explored by leveraging pre-trained LLMs for synthetic data generation \cite{moslem-etal-2022-domain}. This approach addresses two scenarios: one with limited bilingual in-domain data and another with no bilingual in-domain data. In the former case, target-side synthetic text is generated using GPT-J \cite{gpt-j} (English) and mGPT \cite{shliazhko-etal-2024-mgpt} (Arabic), followed by back-translation \cite{sennrich-etal-2016-improving} to create parallel data. In the latter case, source text is forward-translated using an MT model before applying the same augmentation technique. A mixed fine-tuning strategy with in-domain data oversampling is employed to ensure balanced learning, resulting in significant BLEU \cite{papineni-etal-2002-bleu,post-2018-call} score improvements of approximately 5-6 for Arabic-English and 2-3 for English-Arabic. The improvement in quality has been also confirmed by manual evaluations.

Another synthetic data generation technique, LexMatcher, focuses on addressing missing senses of polysemous words \cite{yin-etal-2024-lexmatcher}. By using bilingual dictionaries, this method retrieves data from parallel corpora to include existing senses. Subsequently, data augmentation is performed with an LLM to generate new sentence pairs for uncovered word senses, enhancing translation disambiguation and sense coverage.
\begin{tcolorbox}[colback=yellow!20,colframe=black!75]
LLM-based synthetic data generation is an effective strategy for domain adaptation and low-resource MT, improving performance primarily by increasing parallel coverage and diversity rather than through explicit reasoning signals. While techniques such as back-translation, sense-aware augmentation, and large-scale forward translation can rival or surpass real-data training when resources are scarce, their effectiveness is constrained by hallucinations, adequacy errors, and domain mismatch. Recent work shows that carefully filtered synthetic preference data can further improve robustness via preference-based training, highlighting the importance of quality control over the generative process itself.\end{tcolorbox}
The use of word and phrase alignment for synthetic data generation has also been studied \cite{dabre-etal-2024-effective}. However, findings indicate that synthetic data produced by LLMs does not offer significant improvements over rule-based and neural-based approaches \cite{frontull-moser-2024-rule}, although the study focused on only one language. 
On a broader study, covering more languages, the authors in \cite{degibert2025scalinglowresourcemtsynthetic} generate a document-level synthetic corpus by forward-translating English Europarl data into seven diverse low-resource languages, later expanding it to 147 additional language pairs via pivoting. They evaluate the quality of this data using automatic metrics and human annotation, confirming its general reliability. Crucially, they show that MT models trained solely on this synthetic data can rival or outperform baselines trained on real data, and that fine-tuning strong pretrained systems like NLLB \cite{nllbteam2022languageleftbehindscaling} and LLaMA-3B \cite{grattafiori2024llama3herdmodels} on this data yields significant performance improvements. Furthermore, their experiments highlight that even noisy synthetic data can be highly effective when clean alternatives are scarce.
A large-scale study examining the use of LLM-generated reasoning traces and CoT supervision for MT, found that explicit reasoning signals do not improve translation quality \cite{zebaze2025llmreasoningmachinetranslation}. Instead, the gains attributed to LLM-based augmentation arise primarily from the inclusion of additional translation attempts rather than abstract reasoning content. This suggests that the effectiveness of LLM-generated synthetic data depends on its contribution to parallel signal and coverage, rather than on the presence of intermediate reasoning or explanation tokens.
When LLMs are used to generate synthetic parallel data for low-resource machine translation, the primary quality challenges are hallucination, adequacy errors, bias, and domain mismatch. Hallucinations are significantly more frequent in low-resource directions, where the model has weaker linguistic grounding, and they often persist even after fine-tuning, effectively teaching the MT system to ignore or distort source content \cite{guerreiro-etal-2023-hallucinations,de-gibert-etal-2025-scaling}.
A second major issue concerns diversity : both lexical and structural. Deterministic decoding strategies tend to produce homogeneous translations, reducing lexical variety. This lack of diversity weakens the regularization effect of synthetic data and can cause fine-tuned models to overfit to narrow phrasing styles. Empirical work shows that more diverse synthetic data improves downstream MT performance \cite{chen2024diversitysyntheticdataimpact}. However, increasing diversity without quality control risks amplifying noise, creating a trade-off between coverage and correctness \cite{iyer-etal-2024-quality}.

Recent work also explores leveraging LLMs to construct synthetic preference signals where they generate multiple candidate translations using LLMs and automatically derive preference pairs through heuristic filtering and quality estimation with COMET, without relying on human annotations \cite{vajda-etal-2025-improving}. These preference pairs are then used to fine-tune MT models via DPO improving translation robustness and reducing structural errors such as truncation and formatting inconsistencies. Their results demonstrate that LLM-generated preference data can effectively complement parallel-data-based augmentation, particularly in low-resource settings. 
Complementary to synthetic data and preference-based supervision, retrieval-based approaches show that target-side monolingual data can be leveraged effectively without explicit data generation or retraining \cite{reheman-etal-2024-exploiting}. By pairing source inputs with semantically similar target-language sentences and integrating them through non-parametric decoding, these methods achieve strong domain adaptation gains, suggesting that much of the benefit attributed to LLM-based augmentation arises from increased target-side coverage rather than from the generative process itself.
Treating preference data as a design variable rather than a byproduct of model diversity helps alleviate several practical bottlenecks in low-resource RLHF, including limited annotation budgets, unstable reward models trained on sparse data, and heavy reliance on biased automatic proxies. This view is consistent with recent findings showing that carefully constructed synthetic and weakly supervised preference signals can enable effective RLHF for low-resource languages without large-scale human annotation \cite{dang2024rlhfspeaklanguagesunlocking}. It also aligns with evidence that misalignment between model behavior and human linguistic preferences is amplified in low-resource settings, making the structure, calibration, and difficulty of preference data especially important \cite{Pava2025MindLanguageGap}. Together, these results indicate that the primary obstacle to RLHF in low-resource machine translation lies not in the choice of optimization objective, but in the challenge of acquiring well-structured, informative, and human-aligned preference signals at scale.

\section{Translation Methodologies}
% MT using LLMs can be broadly categorized into two primary methodological approaches: (1) prompting-based techniques and (2) fine-tuning-based techniques. The former leverages pre-trained models with minimal adaptation, whereas the latter involves modifying model parameters to enhance performance on translation tasks. The methods can be combined - prompting based techniques can be used after fine-tuning. In this section, we discuss these methodologies in detail.

MT using LLMs can be broadly categorized into two primary methodological approaches: (1) \emph{prompting-based techniques} and (2) \emph{fine-tuning-based techniques}. Prompting-based methods leverage pre-trained LLMs with no parameter updates, eliciting translation behavior through carefully designed instructions and in-context examples. In contrast, fine-tuning-based methods explicitly modify model parameters to improve translation performance, often obtaining stronger and more stable results at the cost of additional computation. These two paradigms are not mutually exclusive: prompting-based techniques are frequently applied after fine-tuning to further steer model behavior at inference time. In this section, we discuss both methodologies in detail. We provide a simplistic guide of the recommended methods to use under different setups in \autoref{tab:which-method-when}. Note that these are general trends that we observed from the paper and there are many contradicting observations, therefore we recommend additional study for specific use-cases.
\begin{table}[t]
\centering
\setlength{\tabcolsep}{5pt}
\renewcommand{\arraystretch}{1.3}

\resizebox{\textwidth}{!}{%
\begin{tabular}{
p{3.2cm}
p{3.2cm}
p{6.5cm}
p{3.5cm}
}
\toprule
\textbf{MT Scenario / Constraint} &
\textbf{Best-Suited Approach} &
\textbf{Why It Works Well} &
\textbf{Key Limitations} \\
\midrule

High-resource language pairs &
Prompting or light PEFT (LoRA) &
Strong multilingual pretraining; minimal adaptation required &
Over-fine-tuning may reduce stylistic control and generalization \\

Mid-resource languages &
PEFT (LoRA / QLoRA) with mixed data &
Balances efficiency and specialization; stable gains &
Direction-dependent performance (En-to-XX harder) \\

Low-resource but related languages &
Few-shot ICL plus synthetic data &
Cross-lingual transfer via shared scripts and morphology &
Sensitive to hallucination and synthetic noise \\

Extremely low-resource languages &
Encoder--decoder MT with explicit supervision &
Parallel or synthetic signal dominates over prompting, encoder-decoder to be preferred &
LLM prompting alone underperforms \\

Rapid domain adaptation (no retraining) &
Dictionary or RAG-based prompting &
Terminology grounding at inference time &
Weak document-level consistency \\

High-stakes domains (medical, finance) &
PEFT or full SFT with terminology constraints &
Parameter-level grounding improves factual fidelity &
Requires curated data; higher cost \\

Document-level terminology consistency &
Fine-tuning with long-context tracking &
Maintains global lexical coherence &
Still unsolved at scale \\

Literary translation &
Long-context prompting with refinement &
Captures style, discourse, and narrative flow &
Automatic metrics under-represent gains \\

Instruction-controlled MT (gender, formality) &
Instruction-tuned LLMs with prompting &
Natural-language constraints at inference time &
Constraint interactions are fragile \\

User preference alignment &
Preference-based fine-tuning (CPO, RLHF) &
Aligns outputs with subjective preferences &
Preference data scarce in low-resource settings \\

Stable, scalable deployment &
Hybrid LLM encoder with NMT decoder &
Combines flexibility and efficiency &
Increased system complexity \\

Automatic MT evaluation (system-level) &
LLM-as-judge &
High correlation after aggregation &
Segment-level instability \\

Automatic MT evaluation (segment-level) &
Learned metrics (COMET, xCOMET) &
More stable fine-grained scoring &
Less interpretable \\

Error analysis and interpretability &
Structured LLM-based evaluation &
Human-readable explanations &
Computationally expensive; prompt-sensitive \\

\bottomrule
\end{tabular}%
}
\caption{Scenario-based recommendations for LLM-based MT; These recommendations reflect observed trends in the literature. However, evidence is sometimes conflicting, and several recommendations are based on a limited number of studies. Readers are therefore cautioned against relying on this table as definitive guidance }
\label{tab:which-method-when}
\end{table}

\paragraph{Pre-trained vs. Instruction-Tuned LLMs.}
An important practical distinction underlying these methodologies concerns whether the LLM is \emph{pre-trained} (base) or \emph{instruction-tuned}. Pre-trained LLMs are optimized solely via next-token prediction and do not explicitly learn to follow natural language instructions. As a result, prompting-based techniques discussed in \autoref{subsec:prompting} are primarily designed for such models, relying on prompt templates and in-context examples to induce translation behavior. However, it is to be noted that pre-trained models often fail to stop at the end of the translation and keeps generating longer sequences. Therefore, often good post-processing strategy is required for pre-trained models.

Instruction-tuned LLMs, on the other hand, are further aligned to follow instructions, maintain roles, and execute multi-step reasoning. The advanced prompting strategies and agentic workflows described in \autoref{subsec:advanced-prompting} including chain-of-thought prompting, self-refinement, and multi-agent debate critically depend on this instruction-following capability. While these techniques are not theoretically restricted to instruction-tuned models, empirical evidence suggests that their stability and effectiveness are substantially higher when applied to instruction-aligned LLMs. It has been observed that instruction tuning and reduce in-context learning capabilities of a model, albeit the study was not specific for MT \cite{wang2024on}.

Fine-tuning-based methods (\autoref{subsec:finetuning}) can be applied to both pre-trained and instruction-tuned LLMs. However, fine-tuning often shifts model behavior toward task-specific optimization, potentially narrowing general instruction-following abilities. Consequently, prompting and fine-tuning should be viewed as complementary tools rather than competing alternatives when designing LLM-based MT systems.

\subsection{Prompting-Based Techniques}
\label{subsec:prompting}
Prompting leverages pre-trained LLMs by conditioning them on task-specific instructions or examples without modifying their underlying parameters. This approach is particularly useful when computational resources for fine-tuning are limited. We present short summary of selected surveyed methods in \autoref{tab:prompting_mt}.
\subsubsection{Zero-Shot Prompting}
In zero-shot prompting, the LLM is provided with an instruction to translate a sentence without any examples. The model relies solely on its pre-trained knowledge to generate translations.
\begin{equation}
\hat{Y} = f_{\theta_{LLM}}(\text{prompt}, X)
\end{equation}

\begin{figure}[h]
    \centering
    \includegraphics[width=\linewidth]{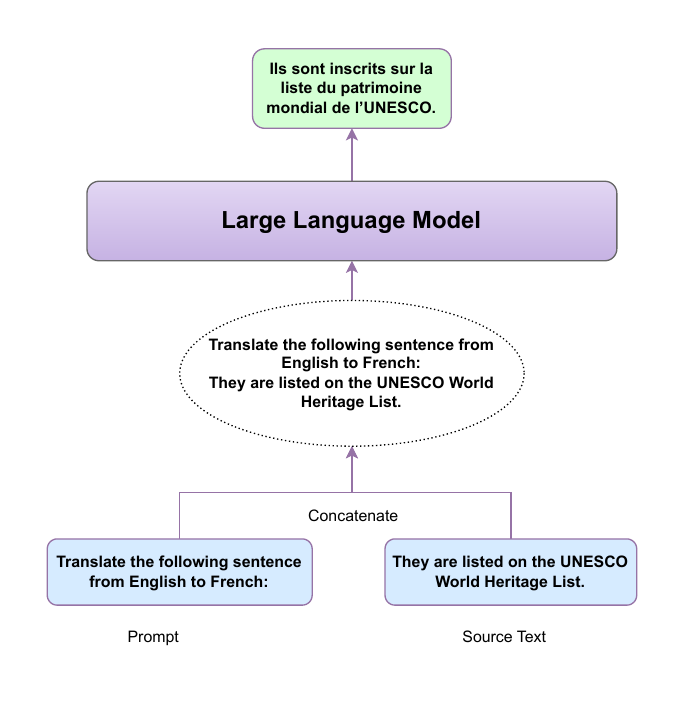}
    \caption{In zero-shot learning, the instruction or prompt is combined with the input text and fed into the LLM. It may also include additional details such as domain-specific information, keywords, or even non-parallel examples.}
    \label{fig:zero-shot}
\end{figure}
Different approaches in zero-shot prompting vary by prompt format and the use of a pivot language for low-resource languages. It has been observed that zero-shot prompting with ChatGPT lags behind MT systems by Google MT \cite{wu2016googlesneuralmachinetranslation}, Tencent \cite{jiao-etal-2022-tencents}, and DeepL by around 5.0 BLEU points \cite{jiao2023chatgptgoodtranslatoryes}. Pivot prompting has been explored to translate between distant languages, where the LLM first translates the sentence to English and then into the target language. This strategy, which uses a resource-rich language (English) as a pivot, improves translation quality between De$\rightarrow$Zh and Ro$\rightarrow$Zh \cite{jiao2023chatgptgoodtranslatoryes}. Human evaluation on the generated translations reveals that GPT-4 \cite{openai2024gpt4technicalreport} makes fewer mistakes compared to ChatGPT and Google MT. Interestingly, although GPT-4 scores lower in BLEU compared to Google MT, human annotators rank its translations more favorably in blind evaluations \cite{jiao2023chatgptgoodtranslatoryes}. 
\begin{align}
\hat{Y}_{\text{pivot}} &= f_{\theta_{LLM}}(\text{prompt}_{L_1 \rightarrow L_{\text{pivot}}}, X) \\
\hat{Y} &= f_{\theta_{LLM}}(\text{prompt}_{\text{pivot} \rightarrow L_2}, \hat{Y}_{\text{pivot}})
\end{align}

To further enhance zero-shot translation, additional information such as domain-specific context and keywords has been incorporated into the prompts. Including domain information leads to improved translation quality compared to vanilla zero-shot prompting, while providing around 10 keywords related to the domain achieves even better COMET scores \cite{aycock-bawden-2024-topic,10806871}. Similarly, adding POS tags in the prompt has been shown to generate slightly better quality translations \cite{10.1145/3700410.3702123}.

\textbf{Variations in Prompt Template}

Prompts that specify only the source and target language have been found to achieve the best overall results, using a simple format \cite{10.5555/3618408.3618846} such as:  
\begin{verbatim}
    [src_lang]: input  
    [tgt_lang]:  
\end{verbatim}  
Similarly, the authors in \cite{jiao2023chatgptgoodtranslatoryes} tested three prompt variations and the following format was found to yield the best performance:  
\begin{verbatim}
    Please provide the [tgt_lang]  
    translation for these sentences:  
\end{verbatim}   

Using pre-trained models for translation without instruction tuning poses challenges, as the model often generates responses that are diverse and difficult to parse. For instance, it sometimes continues generating tokens even after completing the translation \cite{alves-etal-2023-steering,bawden-yvon-2023-investigating}. A common solution is to discard tokens after a newline; however, models frequently produce outputs in patterns that do not match known criteria, making this strategy unreliable. These issues are generally mitigated when models undergo fine-tuning \cite{alves-etal-2023-steering,bawden-yvon-2023-investigating}.

\subsubsection{In-context Learning}

In-Context Learning (ICL) enhances MT by providing a model with a small set of input-output examples before generating a translation, helping it infer correct patterns. This method significantly improves translation accuracy, especially for domain-specific or idiomatic expressions, by guiding the model toward appropriate mappings. The effectiveness of the approach depends on multiple parameters, including \textit{the number of in-context examples, selection of the examples, and their ordering}.

\begin{figure*}[h]
    \centering
    \includegraphics[width=0.8\linewidth]{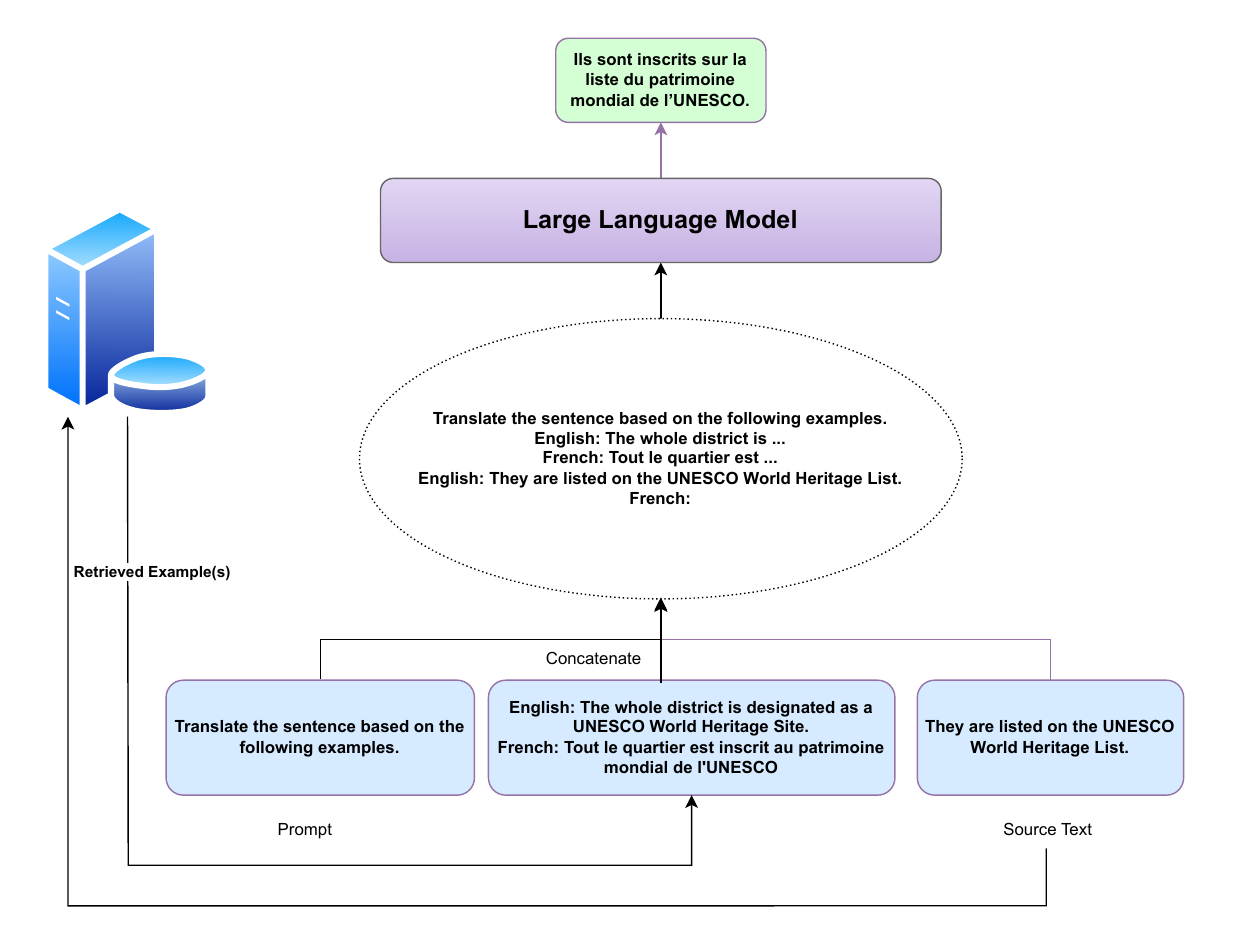}
    \caption{In in-context learning, the model is provided with a few examples (input-output pairs) along with the instruction or prompt. These examples help the LLM infer the desired pattern and generate appropriate responses.}
    \label{fig:icl}
\end{figure*}

\begin{equation}
\hat{Y} = f_{\theta_{LLM}}(\text{prompt}, \{(X_i, Y_i)\}_{i=1}^{n}, X)
\end{equation}
However, the use of in-context examples does not always result in better output quality. It has been observed that in some cases, few-shot prompting can lead to poorer performance compared to one-shot prompting \cite{alves-etal-2023-steering}. When models are fine-tuned without few-shot examples, they exhibit a higher likelihood of hallucination when prompted with few-shot examples during inference. Fine-tuning with few-shot examples has been shown to mitigate this issue, reducing the chances of hallucination and enhancing output quality.

\textbf{Number of In-Context Examples:}
One critical decision when designing prompts for LLMs is determining the optimal number of in-context examples. Increasing the number of examples can enhance performance, but it also comes with higher computational costs and diminishing returns. Using too few examples may bias the model towards the style of the provided instances, while incorporating too many may not lead to proportional improvements in output quality.

Several studies have explored this trade-off. Research has shown consistent improvements in COMET scores when increasing from one to five in-context examples \cite{hendy2023goodgptmodelsmachine}. Similarly, a significant rise in BLEU scores has been reported when increasing from a single example to five, although results for intermediate number of examples were not provided \cite{mu-etal-2023-augmenting}.

A more detailed investigation assessed performance with 0, 1, 2, 5, and 10 in-context examples. The findings indicate that additional examples generally improve results, but the gains diminish beyond five-shot prompting \cite{vilar-etal-2023-prompting}. This aligns with previous research suggesting that high-quality in-context examples can yield translation performance comparable to fine-tuned models {\cite{10.5555/3618408.3618846}. This is particularly useful when fine-tuning is infeasible due to computational limitations or unavailable model weights. Other studies observed significant improvements when increasing from 1-shot to 5-shot prompting, although gains were less pronounced for 10-shot, with some cases where 5-shot outperformed both 10-shot and 20-shot \cite{10.5555/3618408.3620130}. Further experiments with 1, 2, and 3-shot demonstrations showed the best results with 3-shot examples on the ChatGLM3-6B model \cite{glm2024chatglm,tang2025adaptivefewshotpromptingmachine}.

These findings collectively suggest that while increasing the number of in-context examples enhances performance, there exists a threshold beyond which further improvements are marginal.
\begin{tcolorbox}[colback=yellow!20,colframe=black!75]
Overall, while increasing the number of examples can enhance LLM performance, studies suggest that five-shot prompting often strikes a balance between efficiency and effectiveness, with diminishing returns beyond this point.
\end{tcolorbox}

\textbf{Selection of In-Context Examples:} 
Selecting effective in-context examples is crucial for optimizing translation performance in LLMs. While strategic selection of examples can enhance results, studies suggest that random selection may sometimes be just as effective, provided the examples are of high quality \cite{vilar-etal-2023-prompting}. The overall quality of examples, regardless of whether they are randomly or carefully chosen, plays a significant role in performance.

The impact of noisy or irrelevant examples has also been investigated. Research indicates that even a single poor-quality example can severely degrade translation quality, particularly in 1-shot scenarios \cite{agrawal-etal-2023-context}. Increasing the number of examples can mitigate the negative effects of one bad sample, but optimal performance is achieved when selecting high-quality, relevant exemplars. Notably, a single optimized example has been found to outperform multiple random ones. Additionally, semantic similarity between the input and in-context examples has been positively correlated with translation quality \cite{10.5555/3618408.3620130}.

One method for optimizing example selection is CTQScorer, which refines prompt selection by incorporating multiple linguistic and structural features \cite{m-etal-2023-ctqscorer}. This approach starts with BM25 retrieval to generate a candidate pool and then ranks examples based on factors such as semantic similarity, translation quality, sentence length, and perplexity. By leveraging a contextual translation quality score, this method has demonstrated a 2.5-point improvement in COMET score over random selection.

Further advancing example selection strategies, research has highlighted the importance of coherence in in-context learning \cite{sia-duh-2023-context}. Rather than relying solely on similarity-based retrieval, coherence is defined as the semantic and stylistic consistency between prompt examples and the test sentence. Studies show that ensuring coherence, aligning prompts in meaning, style, and contextual flow have better translation performance than selecting examples purely based on word or embedding similarity. This approach shifts the focus from isolated, highly similar examples to logically consistent prompt setups, redefining how in-context examples should be chosen for MT.
\begin{tcolorbox}[colback=yellow!20,colframe=black!75]
Carefully selecting high-quality, relevant in-context examples significantly improves translation performance, while even one bad example can degrade results. Beyond similarity, factors like coherence, diversity, and structured selection (e.g., submodular functions, CTQScorer) enhance few-shot translation. In low-resource settings, source-similar or cross-lingual examples can sometimes outperform same-language examples, leveraging LLMs' transfer learning capabilities.\end{tcolorbox}
An alternative approach involves using submodular functions to optimize example selection \cite{ji-etal-2024-submodular}. This technique ensures that adding a new example improves coverage while avoiding excessive redundancy. By balancing content overlap, target alignment, and diversity, submodular-selected prompts have consistently outperformed random and purely similarity-based selections across multiple datasets, leading to superior few-shot translation results.

In low-resource scenarios, selecting source-similar examples has been found to yield better BLEU and COMET scores than using arbitrary examples \cite{zebaze2024incontextexampleselectionsimilarity}. When data is scarce, closely matched examples can provide valuable lexical and structural guidance that the model might not otherwise recall. Additionally, the makeup of the candidate pool influences selection effectiveness. When limited to high-quality reference translations, the performance gap between random and similarity-based selection decreases.

The authors in \cite{mu-etal-2023-augmenting} used translation memories (TMs) to enhance LLMs translation performance. The approach retrieves the top-k most similar sentence pairs from a TM database and incorporates them into prompts using structured templates. Two main template styles, instruction-style and code-style are used to provide translation hints. Experiments demonstrate that TM-based prompting significantly improves LLM translation quality, often surpassing strong NMT baselines.

The importance of correctly pairing in-context examples has also been emphasized \cite{zhu-etal-2024-multilingual}. Mismatched examples, such as incorrect translations, can lead to complete translation failures. Interestingly, cross-lingual demonstrations have been shown to be surprisingly effective in low-resource translation scenarios. Using examples from a different but better-known language pair has, in some cases, outperformed using same-language examples. This suggests that LLMs may leverage general translation patterns across linguistically related languages, improving translation quality even when direct examples in the target language are unavailable.
Fragment-Shot Prompting \cite{frontull2025compensatingdatareasoninglowresource}, a retrieval-augmented in-context learning strategy that selects examples based on syntactic coverage rather than sentence-level semantic similarity. Instead of retrieving full sentences, the method decomposes the input into fragments and retrieves parallel examples that explicitly contain these fragments, ensuring broad lexical and structural coverage in the prompt. Their results demonstrate that syntactic coverage correlates strongly with translation quality when translating into or between low-resource languages, while offering limited gains for low-to-high resource directions. 

Collectively, these findings underscore the importance of intelligent example selection considering factors such as relevance, quality, coherence, and diversity to maximize few-shot translation performance.

\textbf{Example Ordering:}

The ordering of examples in the prompt dataset plays a important role in determining the quality of translations. This section explores how dataset ordering can be optimized to enhance translation performance. Using a BERT-based re-ranker model to rank the examples has been shown to improve BLEU \cite{papineni-etal-2002-bleu}, ROUGE \cite{lin-2004-rouge}, CHRF \cite{popovic-2015-chrf}, and COMET \cite{rei-etal-2020-comet} scores \cite{tang2025adaptivefewshotpromptingmachine}. Similarly, examining the effect of example order in few-shot prompts reveals that task-level prompts are less sensitive to ordering compared to randomly sampled examples, resulting in lower standard deviations and better translation quality. Testing all permutations of four randomly sampled and top task-level examples demonstrates that the ordering of in-context examples significantly impacts MT performance, as indicated by notable BLEU variations across three runs with random examples \cite{agrawal-etal-2023-context}. 
\begin{tcolorbox}[colback=yellow!20,colframe=black!75]
The ordering of examples in a prompt can influence the quality of model outputs. Various methods exist for ranking candidate examples; however, placing the most similar or highest-ranked examples closer to the current input is a common strategy. This approach generally yields slightly better results compared to alternative ordering techniques.
\end{tcolorbox}
Experiments with two types of ordering, least similar to most similar (followed by the input sentence) and most similar to least similar (followed by the input sentence), show no significant improvements of either method over the other. However, the former yields slightly higher average BLEU and COMET scores \cite{zebaze2024incontextexampleselectionsimilarity}.

\subsection{Advanced Prompting}
\label{subsec:advanced-prompting}

\begin{table}[t]
\centering
\small
\setlength{\tabcolsep}{6pt}
\renewcommand{\arraystretch}{1.25}
\resizebox{\textwidth}{!}{%
\begin{tabular}{p{3cm} p{2cm} p{3.8cm} p{7cm}}
\toprule
\textbf{Paper} &
\textbf{Model(s)} &
\textbf{Prompting Strategy} &
\textbf{Key Observation} \\
\midrule

Jiao et al. (2023) \cite{jiao2023chatgptgoodtranslatoryes} 
& ChatGPT, GPT-4 
& Zero-shot / Pivot Prompting 
& GPT-4 approaches commercial MT quality in high-resource settings, though BLEU remains lower than specialized MT systems. \\

Alves et al. (2023) \cite{alves-etal-2023-steering} 
& GPT-3, BLOOM 
& Few-shot Prompting 
& Prompting alone produces unstable outputs and hallucinations; minimal fine-tuning improves controllability. \\

Vilar et al. (2023) \cite{vilar-etal-2023-prompting} 
& GPT-3 
& In-context Learning 
& Translation quality improves with additional examples but saturates around five-shot prompting. \\

Mu et al. (2023) \cite{mu-etal-2023-augmenting} 
& GPT-3.5 
& TM-augmented Prompting 
& Injecting translation memories into prompts significantly improves LLM translation quality, often surpassing NMT baselines. \\

Peng et al. (2023) \cite{peng-etal-2023-towards} 
& ChatGPT 
& Chain-of-Thought Prompting 
& Step-by-step reasoning degrades translation fluency and adequacy, showing misalignment between CoT and MT. \\

Lu et al. (2024) \cite{lu-etal-2024-chain} 
& GPT-3.5, GPT-4 
& Chain-of-Dictionary Prompting 
& Lexical chains via auxiliary languages improve low-resource translation without parallel data. \\

Puduppully et al. (2023) \cite{puduppully-etal-2023-decomt} 
& mT5 
& Decomposed Prompting 
& Chunk-wise translation with contextual refinement improves low-resource and long-sentence translation quality. \\

Chen et al. (2024) \cite{chen-etal-2024-iterative} 
& GPT-3.5, GPT-4 
& Iterative Refinement 
& Self-refinement improves fluency and human preference despite slight degradation in BLEU. \\

He et al. (2024) \cite{he-etal-2024-exploring} 
& GPT-3.5 
& Multi-Aspect Prompting 
& Explicit modeling of keywords, topics, and examples reduces hallucination and ambiguity. \\

Feng et al. (2025) \cite{feng-etal-2025-tear} 
& GPT-4 
& Self-Refinement (TEaR) 
& LLM-driven error detection and refinement reduces MQM errors without external supervision. \\

Liang et al. (2024) \cite{liang-etal-2024-encouraging} 
& GPT-3.5 
& Multi-Agent Debate 
& Debate-based prompting mitigates self-confirmation bias and improves translation accuracy. \\

Zebaze et al. (2025) \cite{zebaze2025compositionaltranslationnovelllmbased} 
& GPT-3.5 
& Compositional Prompting 
& Phrase-level decomposition followed by recomposition improves low-resource and out-of-domain translation. \\

\bottomrule
\end{tabular}%
}
\caption{Prompting-based LLM methods for machine translation.}
\label{tab:prompting_mt}
\end{table}

\subsubsection{Chain-of-Thought based Methods}
From a Chain-of-Thought (CoT) perspective, one of the early studies investigated whether step-by-step reasoning prompts, successful in many NLP tasks, can enhance ChatGPT\'s performance in MT \cite{peng-etal-2023-towards}. They apply both zero-shot and 1-shot CoT prompting, where ChatGPT is instructed to translate source sentences word-by-word before generating the full translation. However, unlike in reasoning tasks, CoT induces a literal, phrase-by-phrase translation behavior that hampers fluency and semantic coherence, leading to a significant drop in COMET scores (e.g., -8.8 for EN$\rightarrow$ZH). This degradation highlights a misalignment between CoT prompting and the holistic nature of translation, suggesting that while CoT can benefit structured reasoning, it is ill-suited for MT directly.

Chain-of-Dictionary Prompting (COD) \cite{lu-etal-2024-chain} is a prompting strategy designed to enhance the multilingual translation capabilities of LLMs, especially in low-resource settings. Rather than relying on few-shot in-context examples which may be difficult to curate or ineffective for rare languages. COD embeds chained lexical mappings directly into the prompt. These chains consist of a keyword from the source sentence and its translations through multiple languages, typically progressing from the target language to high-resource auxiliary languages (e.g., French, German, Portuguese). Formally, the prompt includes two parts: a standard instruction and a sequence of dictionary hints like \textit{river means {\dn ndF }  means  fleuve  means  Fluss}.  These lexicon chains act as contextual anchors, helping the model to associate meaning more accurately across languages. These hints are prepended to the translation prompt, thus constraining the model's lexical space and guiding its cross-lingual reasoning. Empirical evaluations demonstrate that this strategy yields significant improvements on the FLORES-200 benchmark, particularly for low-resource languages, and even outperforms traditional few-shot and dictionary-based prompting baselines.
Recognizing the monotonic alignment typically found in related languages that share syntactic and lexical similarities, \cite{puduppully-etal-2023-decomt} propose DecoMT, a decomposed prompting approach for few-shot MT. DecoMT breaks down the translation process into two stages. In the first stage, the input sentence is segmented into small chunks, each of which is translated independently using few-shot prompting with an encoder-decoder language model (mT5). In the second stage, these preliminary translations are refined through contextual translation, which involves infilling masked tokens in the context of adjacent chunks and previously predicted outputs. This incremental process leverages the bidirectional capacity of mT5 to produce more coherent and contextually appropriate translations. DecoMT is especially effective for low-resource language pairs, demonstrating substantial improvements over established few-shot prompting baselines - achieving up to 8 chrF++ points gain on average across evaluated languages. The method also proves robust for longer sentences and maintains competitive performance in high-resource settings, offering a practical solution without retraining large models, albeit with a modest requirement for manually aligned prompt examples.
The authors in \cite{chen-etal-2024-iterative} proposed a prompting-based approach where an LLM iteratively refines a translation by repeatedly conditioning on the source sentence and the previous output. They repeat the cycle for 4 iterations and best iteration chosen post-hoc using COMETQE. Although iterative refinement often reduces BLEU due to increased lexical diversity, neural metrics and human evaluations show improvements in fluency and naturalness. 

The MAPS (Multi-Aspect Prompting and Selection) \cite{he-etal-2024-exploring} framework introduces a prompting strategy that mimics human translation behavior by guiding LLMs through a structured, three-step process. First, the LLM is prompted to extract translation-relevant knowledge from the source sentence in three forms: keywords, the underlying topic, and relevant example translations using manually designed few-shot prompts. Next, each type of extracted knowledge is separately reintegrated into new prompts to guide the LLM in generating multiple translation candidates. Finally, a selection mechanism, either via a reference-free quality estimation model (COMET-QE) or prompting the LLM itself (LLM-SCQ), is used to choose the best candidate. This multi-faceted prompting strategy improves translation quality across various language pairs and also helps to reduce common errors such as hallucination, ambiguity, and mistranslation without relying on external data or curated in-context examples.

Another method which use 
Recognizing that translations generated by LLMs often contain subtle errors despite strong benchmark performance, the authors in \cite{feng-etal-2025-tear} propose TEaR - a plug-and-play self-refinement framework for MT that improves translation quality through internal feedback loops. TEaR operates in three stages: \textit{Translate}, where an LLM produces an initial translation; \textit{Estimate}, where the same LLM simulates human-like error assessment using MQM guidelines; and \textit{Refine}, where the LLM revises the translation based on its self-generated evaluation. This self-contained design allows TEaR to bypass reliance on external models or human feedback. The Estimate module identifies major and minor translation errors such as mistranslation, awkward style, and grammar issues and only proceeds to refinement if necessary, making the approach more selective and stable than baselines like CoT. Experiments on WMT22 and WMT23 datasets across eight translation directions show that TEaR consistently improves translation quality and reducing MQM error counts more effectively than other self-refinement or external feedback systems. Moreover, iterative application of TEaR further enhances translation quality in successive rounds, demonstrating its capacity for continuous improvement.

While prior works attempt to mimic human cognitive behaviors in prompting such as self-reflection, iterative refinement, or chain-of-thought reasoning to improve model performance, these methods often fall short when it comes to generating truly divergent or corrective reasoning paths. In particular, self-reflection-based approaches can suffer from what the authors term the Degeneration-of-Thought (DoT) problem: once an LLM gains confidence in its initial (possibly incorrect) answer, it becomes resistant to revising or challenging that stance in subsequent iterations. To address this, the paper proposes the Multi-Agent Debate (MAD) framework \cite{liang-etal-2024-encouraging}, which draws on the human problem-solving strategy of adversarial discussion. MAD simulates a structured debate between multiple LLM agents - typically an affirmative and a negative debater, who challenge each other in a \textit{tit for tat} style, with a judge model overseeing the debate and determining the final answer.

This debate setup introduces external feedback and perspective shifts that self-reflection lacks, promoting more diverse and accurate reasoning. The MAD framework is evaluated on two challenging tasks: Commonsense MT and Counter-Intuitive Arithmetic Reasoning where it significantly outperforms baseline methods including self-reflection and even matches or surpasses stronger models like GPT-4 when used with GPT-3.5 as the base. Key to MAD’s effectiveness are its adaptive break strategy (which ends debates early when a satisfactory solution is found), balanced argumentative intensity, and consistent use of the same LLM for all roles to mitigate judge bias. This methodology not only encourages divergent thinking in LLMs but also offers a scalable, model-agnostic approach to overcoming reasoning limitations in existing prompting strategies.
Compositional Translation (CompTra) \cite{zebaze2025compositionaltranslationnovelllmbased} is a prompting strategy designed to enhance the MT capabilities of LLMs, particularly for low-resource languages. Instead of translating entire sentences directly, CompTra treats translation as a step-by-step reasoning task. First, the input sentence is decomposed into simpler, context-preserving phrases using a prompt that includes examples from datasets like MinWikiSplit. This decomposition is dynamic and structure-aware, producing shorter phrases that are easier for LLMs to handle. Next, each phrase is translated independently in a few-shot setting, where relevant examples are retrieved through similarity search methods like BM25. Incorrectly translated phrases in other languages are filtered using a language identifier to maintain quality. Finally, the LLM is prompted again with the original sentence and the self-generated phrase-translation pairs to produce a complete, coherent translation. This approach takes advantage of the fact that LLMs perform better on shorter inputs and can more easily match relevant in-context examples for them. CompTra consistently outperforms traditional few-shot and zero-shot methods across multiple benchmarks, showing strong improvements in low-resource and out-of-domain scenarios.

\begin{tcolorbox}[colback=yellow!20,colframe=black!75]
Prompting and agentic methods attempt to improve LLM-based MT by decomposing translation into intermediate steps such as lexical grounding, chunk-wise translation, self-refinement, or multi-agent interaction. While naive Chain-of-Thought prompting degrades translation quality by enforcing overly literal reasoning, structured strategies such as dictionary chaining, iterative refinement, and multi-aspect prompting can help. Agentic workflows further enhance robustness through self-critique and staged post-editing, but their effectiveness depends strongly on the quality of intermediate signals and initial translations, with diminishing returns for noisy or low-quality inputs.
\end{tcolorbox}

\subsubsection{Agentic Methods}

Recently, agentic frameworks have gained significant traction in NLP for enhancing reasoning and multi-step problem solving, yet only a few works have examined their applicability to MT. In this context, \citet{grubisic-korencic-2025-irb} introduce a lightweight MT system that pairs an off-the-shelf multilingual LLM (Gemma3-12B) with a simple two-stage \emph{self-refine} workflow, where the model first produces an initial translation and subsequently improves it through prompted self-critique, all without fine-tuning or additional training data. Evaluated across 31 language pairs in the WMT25 shared tasks, their system consistently outperforms the base model, achieves mid-tier rankings overall, and is notably competitive for approximately 16\% of the evaluated language pairs, particularly for mid- and low-resource directions. In contrast, the system is less competitive for high-resource European languages, where many submissions rely on fine-tuned or task-adapted models. Together, these findings demonstrate that even a minimal and resource-efficient agentic setup can offer meaningful performance gains in MT, especially for languages and domains where large-scale supervised adaptation is limited.

Narrowing down the steps within the agentic system,  Multi-agentMT \cite{kim-2025-preliminary, kim-2025-multi} integrates a Translate$\rightarrow$Postedit$\rightarrow$Proofread workflow built on Prompt Chaining and a modified RUBRIC-MQM error-annotation mechanism. They observe that the translation quality depends strongly on both the quality of the initial translation and the explicitness of the Postedit agent's outputs, with the multi-agent pipeline often improving semantic adequacy after the proofread stage. The approach demonstrates promising gains particularly when paired with strong initial translations whereas performance lags when the initial translations are weaker or when error spans are misidentified: highlighting that agentic MT systems are especially competitive for cases where the input quality is moderate-to-high but less effective for structurally noisy or low-quality initial hypotheses.

\subsection{Fine-Tuning Approaches}
\label{subsec:finetuning}

Fine-tuning adapts large language models (LLMs) to machine translation (MT) by updating model parameters using task-specific supervision. Compared to prompting-based methods, fine-tuning typically yields stronger and more stable translation performance, albeit at higher computational cost and with potential side effects on emergent LLM capabilities. Importantly, fine-tuning strategies are not mutually exclusive. Existing approaches can be characterized along three complementary dimensions: \emph{training objectives}, \emph{parameter update strategies}, and \emph{training data composition}. In practice, modern MT systems often combine choices along all three axes, such as parameter-efficient fine-tuning with preference-based objectives over mixed monolingual and parallel data. We provide a short summary of the major used techniques in \autoref{tab:finetuning_mt}. Further, we provide a summary of existing MT-specific LLMs.

\subsubsection{Training Objectives}
\label{subsubsec:training_objectives}

\paragraph{Supervised Fine-Tuning}
Supervised fine-tuning (SFT) optimizes LLMs on parallel corpora by minimizing the cross-entropy loss between predicted translations and gold-standard references. This objective substantially improves translation accuracy, particularly for domain-specific scenarios. SFT has been shown to enable strong multilingual generalization even with limited supervision, where fine-tuning on a single language direction transfers to others \cite{zhu-etal-2024-fine}. However, SFT is sensitive to target-side noise and language imbalance. Using English as the target language can introduce misinterpretations, disproportionately affecting well-represented languages. Effective SFT therefore requires careful alignment to prevent the model from acquiring undesirable biases.

Although SFT consistently improves automatic metrics such as COMET, it can simultaneously degrade several valuable emergent LLM capabilities. Parallel-only fine-tuning weakens formality control, few-shot domain adaptation, and document-level contextualization, with such degradations appearing even when training on relatively small parallel datasets (e.g., 18K examples) and intensifying as data size increases \cite{stap-etal-2024-fine}. These observations suggest that while SFT is effective for improving translation accuracy, it must be applied cautiously to avoid sacrificing broader LLM strengths.

\paragraph{Preference- and Reinforcement-Based Objectives.}
To address the limitations of reference-based supervision, preference-driven and reinforcement learning (RL) objectives have been proposed for MT. Contrastive Preference Optimization (CPO) trains models to distinguish between ``adequate'' and ``near-perfect'' translations using hard negative examples \cite{10.5555/3692070.3694345}. By explicitly rejecting high-quality but imperfect outputs, CPO encourages the generation of more precise translations. CPO has been shown to deliver significant gains even when applied with limited parallel data and minimal parameter updates.

Reinforcement Learning with Human Feedback (RLHF) has also been adapted for MT to align model outputs with human preferences \cite{xu2024advancingtranslationpreferencemodeling}. Instead of relying on expensive human-annotated preference data, high-quality human translations are used as proxies for preferred outputs. The approach involves training a reward model to distinguish human and machine translations, followed by policy optimization using Proximal Policy Optimization (PPO) \cite{schulman2017proximalpolicyoptimizationalgorithms}. This framework improves translation quality and exhibits cross-lingual transfer, generalizing gains to language pairs not explicitly optimized during RL.

More fine-grained RL formulations further refine optimization. Token-level reinforcement learning assigns rewards at the level of individual tokens rather than full sequences \cite{ramos2025finegrainedrewardoptimizationmachine}. Using xCOMET-based error detection \cite{guerreiro-etal-2024-xcomet}, tokens within error spans are weighted according to error severity, enabling continuous and granular feedback. This formulation improves training stability and mitigates overfitting and exposure bias compared to sentence-level RL and SFT baselines. Similarly, MT-R1-Zero introduces a mixed rule–metric reward that combines formatting constraints with automatic MT quality metrics, including BLEU and COMETKiwi \cite{feng-etal-2025-mt-r1}. Optimization with the GRPO algorithm generates strong performance across in-domain, out-of-distribution, and multilingual settings, with most gains attributed to the RL framework itself rather than explicit reasoning prompts.

The authors in \cite{gisserot-boukhlef-etal-2024-preference} analyze CPO as a training objective for LLM-based MT. CPO combines a contrastive preference loss that increases the likelihood gap between preferred and rejected translations with an explicit supervised likelihood term on the chosen output. This formulation removes the need for a reference policy while preserving the ability to encode relative quality distinctions between candidate translations. From a methodological perspective, CPO is well suited to MT, where competing hypotheses often differ subtly in adequacy or fluency rather than exhibiting clear correctness errors. The study positions CPO as a principled objective for translation alignment that operates at the level of optimization, without requiring architectural changes or inference-time decoding modifications. 
Complementary to these approaches, the authors in \cite{zhu-etal-2024-preference} propose a preference-driven fine-tuning paradigm that directly aligns LLM translation probabilities with graded human preferences using a Plackett-Luce formulation. Instead of relying on binary rankings or synthetic negative examples, their method leverages multiple candidate translations per source sentence annotated with fine-grained quality scores, enabling the model to capture nuanced differences in adequacy and fluency. The approach effectively mitigates the performance plateau observed with standard supervised fine-tuning and demonstrates consistent improvements across multiple language directions and evaluation benchmarks, highlighting the value of preference distance information for stable and data-efficient MT optimization. 

Complementary to metric-driven RL, SSR-Zero introduces a self-rewarding reinforcement learning framework that removes dependence on both reference translations and external reward models. In SSR, the same LLM alternates between actor and judge roles: it generates multiple translation candidates and then evaluates them using a referenceless LLM-as-a-judge prompt to assign continuous quality scores, which serve as rewards for GRPO optimization. Trained on only 13K monolingual sentences, SSR-Zero achieves large gains over its base model and surpasses several MT-specialized and larger general-purpose LLMs on English–Chinese benchmarks. Importantly, the study shows that self-generated rewards are competitive with frozen LLM judges and, while slightly weaker than specialized metrics such as COMET, provide complementary benefits when combined with them. 
Overall, this line of work demonstrates that preference-based objectives provide a flexible optimization framework for MT fine-tuning, enabling models to learn relative translation quality directly at the objective level rather than relying solely on reference likelihood maximization. 
\subsubsection{Parameter Update Strategies}
\label{subsubsec:parameter_updates}

\paragraph{Full-Parameter Fine-Tuning.}
Full-parameter fine-tuning updates all model weights and typically achieves strong MT performance, particularly for high-resource language pairs and specialized domains.
From a full SFT perspective, TOWER+ \cite{rei2025towerbridginggeneralitytranslation} uses SFT as a balancing stage rather than a pure MT finetuning step, with translation making up ~22\% of the SFT data and the rest devoted to general instruction-following to preserve broad capabilities. SeamlessM4T \cite{communication2023seamlessmultilingualexpressivestreaming} exemplifies large-scale supervised fine-tuning as the backbone of multilingual text-to-text translation, treating SFT as the primary mechanism for learning translation mappings. It is trained with standard likelihood-based objectives over massive curated and mined parallel corpor and jointly optimized within a multitask framework that includes speech and ASR objectives. It demonstrates that, given sufficient high-quality parallel data and careful multilingual balancing, conventional MLE-based fine-tuning can deliver strong and stable translation performance across dozens of language pairs without reliance on preference modeling or reinforcement learning, while also serving as a strong teacher signal for cross-modal transfer.
Hunyuan-MT \cite{zheng2025hunyuanmttechnicalreport} treats SFT as the primary driver of translation quality, applying large-scale, multi-stage parallel-data SFT (millions of pairs, heavily filtered) before RL refinement.  ALMA \cite{xu2024a} relies entirely on SFT for adaptation, using a minimalist two-stage design: monolingual SFT for multilingual competence followed by small, high-quality parallel SFT for translation induction. GemmaX2 \cite{cui-etal-2025-multilingual} positions SFT as a final specialization layer, applied after extensive continual pretraining, mainly to align the model with translation instructions rather than to teach translation from scratch. However, full fine-tuning is computationally expensive motivating more efficient alternatives.

\paragraph{Parameter-Efficient Fine-Tuning (PEFT).}
PEFT methods aim to reduce computational and memory overhead by updating only a small subset of parameters. Low-Rank Adaptation (LoRA) injects trainable low-rank matrices into frozen pretrained weights, enabling efficient adaptation without modifying the entire model \cite{hu2022lora}. Empirical studies show that adapter-based methods can match full fine-tuning performance while training as little as 10\% of parameters, though effectiveness varies with language distance and dataset size \cite{ustun-cooper-stickland-2022-parameter}. Comprehensive surveys of PEFT methods are provided in \cite{xu2023parameterefficientfinetuningmethodspretrained,han2024parameterefficientfinetuninglargemodels}.

QLoRA further improves efficiency by combining LoRA with quantized model weights \cite{dettmers2023qloraefficientfinetuningquantized}. Applied to Mistral-7B \cite{jiang2023mistral7b}, QLoRA outperforms strong prompting-based baselines when combined with minimal in-context supervision \cite{moslem2023finetuninglargelanguagemodels}. Comparative evaluations show that LoRA-based fine-tuning achieves performance comparable to SFT while updating up to 50$\times$ fewer parameters \cite{alves-etal-2023-steering}, and can surpass few-shot prompting with as few as 2K training examples. Large-scale evaluations across multiple LLMs confirm that QLoRA fine-tuning yields substantial BLEU improvements while training less than 1\% of model parameters \cite{zhang-etal-2023-machine}.

A specialized variant, Language-Specific Fine-Tuning with LoRA (LSFTL), selectively inserts LoRA adapters into attention and feed-forward layers of multilingual MT models \cite{10918960}. This approach enables compact models such as NLLB-200-Distilled-600M to match the performance of substantially larger counterparts while exhibiting faster convergence and improved early-stage performance, making it particularly effective for low-resource language pairs.

\begin{table}
\centering
\small
\setlength{\tabcolsep}{6pt}
\renewcommand{\arraystretch}{1.25}
\resizebox{\textwidth}{!}{%
\begin{tabular}{p{4.2cm} p{3.5cm} p{3.8cm} p{7cm}}
\toprule
\textbf{Paper} &
\textbf{Model(s)} &
\textbf{Fine-Tuning Strategy} &
\textbf{Key Observation} \\
\midrule

Xu et al. (2024) \cite{xu2024a} 
& LLaMA (ALMA) 
& Two-stage SFT 
& Monolingual adaptation followed by parallel fine-tuning yields large multilingual MT gains. \\

Stap et al. (2024) \cite{stap-etal-2024-fine} 
& LLaMA-2 
& Supervised Fine-Tuning 
& Parallel-only fine-tuning improves MT but degrades formality control and contextual generalization. \\

Moslem et al. (2023) \cite{moslem2023finetuninglargelanguagemodels} 
& Mistral-7B 
& QLoRA Fine-Tuning 
& Parameter-efficient fine-tuning outperforms few-shot prompting with limited training data. \\

Alves et al. (2023) \cite{alves-etal-2023-steering} 
& BLOOM 
& LoRA Fine-Tuning 
& LoRA matches full fine-tuning performance while updating 50$\times$ fewer parameters. \\

Zhang et al. (2023) \cite{zhang-etal-2023-machine} 
& Multiple LLMs 
& QLoRA Fine-Tuning 
& Updating less than 1\% of parameters yields substantial BLEU improvements across languages. \\

Gisserot-Boukhlef et al. (2024) \cite{gisserot-boukhlef-etal-2024-preference} 
& ALMA 
& Contrastive Preference Optimization 
& Learning from relative translation quality improves adequacy without reference policies. \\

Xu et al. (2024) \cite{xu2024advancingtranslationpreferencemodeling} 
& LLaMA 
& RLHF 
& Reinforcement learning aligns translations with human preferences and generalizes across language pairs. \\

Ramos et al. (2025) \cite{ramos2025finegrainedrewardoptimizationmachine} 
& LLaMA 
& Token-level RL 
& Fine-grained reward assignment improves training stability and reduces exposure bias. \\

Feng et al. (2025) \cite{feng-etal-2025-mt-r1} 
& LLaMA 
& Rule-Metric RL 
& Mixed rule and metric rewards yield robust multilingual and out-of-domain translation performance. \\

\bottomrule
\end{tabular}%
}
\caption{Fine-tuning-based LLM methods for machine translation.}
\label{tab:finetuning_mt}
\end{table}

\subsubsection{Training Data Composition}
\label{subsubsec:data_composition}

\paragraph{Monolingual and Parallel Data.}
Beyond optimization objectives and parameterization, the composition of training data plays a critical role in fine-tuning outcomes. ALMA introduces a two-stage fine-tuning strategy to address the English-centric bias of many pretrained LLMs \cite{xu2024a}. In the first stage, models are adapted using non-English monolingual data to strengthen multilingual representations. In the second stage, fine-tuning on a relatively small amount of high-quality parallel data aligns the model toward translation generation. This strategy yields large gains over zero-shot baselines across multiple WMT language pairs.

To mitigate the loss of emergent LLM capabilities observed under parallel-only SFT, mixed-data fine-tuning combines monolingual and parallel corpora during adaptation \cite{stap-etal-2024-fine}. Incorporating monolingual data preserves formality control and document-level contextualization while still improving translation quality. These results indicate that maintaining LLM generalization abilities is compatible with strong MT performance when training data and objectives are carefully balanced. Monolingual data, especially data resembling LLM pretraining distributions acts as an effective regularizer, helping retain pre-existing capabilities during task adaptation.

\paragraph{Preference Data}
Beyond the choice of optimization objective, the construction of preference data plays a central role in preference-based fine-tuning for machine translation and implicitly exposes key bottlenecks in applying RLHF to low-resource languages. A central difficulty is the limited availability of high-quality human preference signals, driven by the scarcity of fluent annotators, annotation cost, and challenges in maintaining consistent judgments across dialects, registers, and cultural norms. Consequently, preference datasets in low-resource settings are often small, noisy, or skewed toward extreme quality contrasts, which undermines stable and data-efficient optimization.

Recent work \cite{gisserot-boukhlef-etal-2024-preference} shows that effective preference learning does not require candidate translations from multiple heterogeneous systems, a requirement that would further intensify data collection challenges. Instead, preference pairs constructed from a single model’s own outputs can provide sufficient and more controllable supervision. The critical factor is the relative positioning of chosen and rejected samples: the preferred output should represent a clear improvement over the base translation, while the rejected output should remain within a reasonable quality range. Very low-quality rejected samples offer little useful signal and can exacerbate noise when human feedback is already scarce or weakly supervised.

\subsection{Results on Prompting and Fine-tuning Methods}
We present results for a subset of representative methods surveyed in the literature; however, a large number of existing approaches could not be included due to practical constraints, including substantial complexity in reproducing their codebases and limited computational resources. As a result, our experimental comparison focuses on methods that are both commonly used and feasible to evaluate under a unified experimental setup, allowing for a controlled and interpretable analysis across prompting- and training-based paradigms.

Specifically, we evaluate ICL using few-shot prompting with one, two, and three demonstrations on Qwen2.5-1.5B-Instruct model \cite{qwen2025qwen25technicalreport}. For each shot setting, we compare two demonstration selection strategies: random selection and similarity-based  selection, where demonstrations are retrieved based on semantic similarity between source sentences. This setup allows us to study the effect of both the number of in-context examples and the choice of demonstration selection strategy on translation performance.
We construct similarity-based few-shot evaluation sets by augmenting each test example with in-context demonstration pairs retrieved from the validation split. All data are drawn from the WMT14 English–German and German–English benchmarks \footnote{\url{https://huggingface.co/datasets/wmt/wmt14/viewer/de-en}}. W filter out examples with empty fields and discard sentence pairs shorter than five tokens on either the source or target side.

We compute dense sentence embeddings for all source sentences in the validation set using the pretrained SentenceTransformer model \footnote{\url{https://huggingface.co/sentence-transformers/all-MiniLM-L6-v2}}.  

In addition to prompting-based methods, we also include training-based baselines to contextualize the performance of ICL. We consider full supervised fine-tuning (SFT) on parallel data, as well as parameter-efficient fine-tuning using Low-Rank Adaptation (LoRA).
For training, we fine-tune the Qwen2.5-1.5B-Instruct model for the German--English translation task using both full supervised fine-tuning and parameter-efficient fine-tuning with LoRA. We use randomly selected 100K examples from the WMT14 German--English parallel corpus for training, and use the official validation set used for model selection.

For full supervised fine-tuning, all model parameters are updated during training. Models are trained for 5 epochs with a maximum sequence length of 512 tokens, using a per-device batch size of 16 and gradient accumulation over 8 steps to increase the effective batch size. Optimization is performed with a learning rate of $1\times10^{-4}$ and a cosine learning rate schedule. Mixed-precision training with bfloat16 is used to reduce memory usage. Model checkpoints are evaluated periodically on the validation set, and the best-performing checkpoint is selected based on validation loss.

For LoRA fine-tuning, the base model parameters are kept frozen and low-rank adapters are inserted into all target modules, with a LoRA rank of 8. The remaining training configuration is kept identical to full fine-tuning, including the number of epochs, learning rate, batch size, sequence length, and learning rate schedule. Validation-based checkpoint selection is again performed using validation loss as the criterion.
For evaluation, we disable sampling and use greedy decoding, with a maximum generation length of 256 tokens. All the training and evaluation are done with LLaMA-Factory \cite{zheng-etal-2024-llamafactory} library.
Together, these experiments enable a direct comparison between zero-parameter adaptation via prompting and parameter-updating approaches under comparable data and evaluation conditions.

\begin{table}[t]
\centering
\small
\resizebox{\textwidth}{!}{%
\begin{tabular}{lccccc ccccc}
\toprule
Method & K-shots & Selection 
& \multicolumn{3}{c}{En$\rightarrow$De} 
& \multicolumn{3}{c}{De$\rightarrow$En} \\
\cmidrule(lr){4-6} \cmidrule(lr){7-9}
& & 
& BLEU & chrF & COMET 
& BLEU & chrF & COMET \\
\midrule
1-shot Prompting & 1 & random     & 13.73 & 45.70 & 0.7337 & 26.89 & 55.24 & 0.8202 \\
1-shot Prompting  & 1 & similarity & 12.94 & 45.17 & 0.7149 & 26.61 & 55.30 & 0.8024 \\
2-shot Prompting & 2 & random     & 14.05 & 45.89 & 0.7417 & 26.91 & 55.12 & 0.8206 \\
2-shot Prompting  & 2 & similarity & 14.12 & 46.00 & 0.7385 & 27.10 & 55.23 & 0.8201 \\
3-shot Prompting  & 3 & random     & \textbf{14.42} & 46.20 & \textbf{0.7477} & 27.26 & 55.23 & \textbf{0.8246} \\
3-shot Prompting  & 3 & similarity & 14.34 & \textbf{46.33} & 0.7445 & \textbf{27.28} & \textbf{55.30} & 0.8230 \\
\midrule
Full Fine-tuning  & NA & -- & \textbf{19.42} & \textbf{50.29} & \textbf{0.7941} & 27.78 & 54.18 & 0.8214 \\
LoRA Fine-tuning  & NA & -- & 18.11 & 48.84 & 0.7786 & \textbf{29.07} & \textbf{55.55} & \textbf{0.8350} \\
\bottomrule
\end{tabular}%
}
\caption{Comparison of prompting and fine-tuning strategies for En$\rightarrow$De and De$\rightarrow$En translation.}
\label{tab:mt_results}
\end{table}

From \autoref{tab:mt_results}, across all settings, de$\rightarrow$en consistently achieves higher BLEU, chrF, and COMET scores than en$\rightarrow$de, indicating a strong direction asymmetry. Increasing the number of in-context examples from one to three makes consistent but modest gains. For de$\rightarrow$en, BLEU improves from roughly 26.7 to 27.3 and COMET from about 0.81 to 0.82, while chrF remains largely stable around 55.3. For en$\rightarrow$de, the gains are slightly more pronounced in relative terms, with BLEU increasing from about 13.3 to 14.4 and COMET22 from 0.72 to 0.75. These results suggest that additional shots help, but the improvements quickly saturate as suggested by \cite{vilar-etal-2023-prompting}.

We did noy find any consistent advantage is observed for similarity-based selection over random selection. In one-shot settings, random selection clearly outperforms similarity-based selection in both directions. In two- and three-shot settings, similarity-based selection sometimes generates marginally higher chrF scores. Overall, the differences are small and inconsistent, indicating that similarity-based retrieval does not reliably outperform random sampling in this setup. However, note that the study is limited due to computational constraints and models with larger number of parameters may provide better outcomes with example selection.

Training-based approaches substantially change the performance landscape. For de$\rightarrow$en, full fine-tuning provides only a small improvement over the best few-shot results, while LoRA achieves the best overall performance, with clear gains in BLEU, chrF, and COMET compared to full fine-tuning. In contrast, for en$\rightarrow$de, full fine-tuning yields a large improvement over all few-shot settings (around +5 BLEU and +0.05 COMET), whereas LoRA underperforms relative to full fine-tuning on all metrics.

In summary, few-shot prompting offers limited but consistent gains, yet it cannot match the benefits of training, particularly for the harder en$\rightarrow$de direction. The effectiveness of LoRA appears to be direction-dependent, working very well for de$\rightarrow$en but not for en$\rightarrow$de. We have made the codes publicly available at \url{https://github.com/babangain/llm-mt-survey}.

\subsection{Mixture-of-Experts in MT}
One of the early approaches used sparse MoE \cite{chirkova2024investigatingpotentialsparsemixturesofexperts} for MT. However, the found that width scaling performs better or on par than sparse MoE. 
Mixture-of-Adapters (MoA) \cite{zhang-etal-2024-lightweight} addresses the parameter inefficiency and training instability of sparse MoE by replacing large feed-forward experts with lightweight adapters inserted into each decoder layer. MoA employs a stage-wise training strategy, where language heterogeneity is explicitly modeled via unsupervised clustering and distilled into a supervised gating network, resulting in more stable and meaningful routing decisions. During expert training, a Gumbel-Max routing scheme balances specialization and generalization, preventing expert overfitting. As a result, MoA achieves consistent translation improvements with an order of magnitude fewer parameters than conventional sparse MoE models.
Lingual-SMoE \cite{zhao2024sparse} improves sparse MoE for multilingual MT by explicitly incorporating linguistic hierarchy and language-dependent capacity allocation into the routing process. Instead of token-only routing, it adopts a routing strategy, where a language-level router first selects a subset of experts based on learned target-language representations, followed by token-level expert selection. To handle heterogeneous translation difficulty, Lingual-SMoE further introduces dynamic expert allocation, adaptively adjusting the number of candidate experts per language based on validation performance. This design mitigates over- and under-fitting across low- and high-resource languages, leading to consistent BLEU improvements over dense and vanilla sparse MoE baselines.
Instead of training MoE from scratch, MoE-LLM approach \cite{ZHU2025104078} leverages hybrid transfer learning, combining a shared LLM backbone with sparsely activated experts to capture language-specific and cross-lingual knowledge. The conditional activation of experts enables efficient capacity expansion while maintaining strong generalization, particularly benefiting low-resource and zero-shot translation scenarios. This demonstrates that MoE can be effectively combined with pre-trained LLMs to balance shared representation learning and expert specialization

\subsection{Performance on Specialized Domains}

\begin{tcolorbox}[colback=yellow!20,colframe=black!75]
    In traditional encoder–decoder MT, domain adaptation required explicit fine-tuning, domain-specific parallel data, or complex data selection strategies. LLMs substantially lower this barrier through in-context learning, dictionary-based prompting, and retrieval-augmented generation, enabling domain adaptation at inference time without retraining. In terminology-heavy domains such as IT, law, and medicine, dictionary and RAG-based prompting often achieves large gains at the sentence level. However, domain adaptation is not fully solved. While lexical accuracy improves, LLMs still struggle with domain-specific factual consistency and long-document coherence. Empirical evidence from WMT shared tasks and domain benchmarks shows that parameter-level adaptation remains necessary for reliable deployment in high-stakes domains. Thus, LLMs shift domain adaptation from a data scarcity problem toward a control and consistency problem, especially at document scale.
\end{tcolorbox}
Domain adaptation has long been recognized as a critical requirement in professional settings such as medicine, finance, law, and e-commerce, where models must align with domain-specific language use, factual conventions, and stylistic norms. A central obstacle in this process is the correct handling of domain terminology, which demands lexical precision, consistency, and fidelity to expert knowledge. Prior to the emergence of LLMs, efforts to adapt NMT systems to specialized domains largely focused on enforcing correct terminology usage through architectural or decoding-level interventions. These included hard and soft lexical constraints during beam search \citep{hokamp-liu-2017-lexically,hasler-etal-2018-neural,post-vilar-2018-fast}, training-time adaptations such as modified cross-entropy losses that explicitly bias models toward domain-specific term generation \citep{dinu-etal-2019-training,ailem-etal-2022-encouraging}, and terminology injection or placeholder-based strategies that encode domain knowledge directly into the input or model parameters \citep{dougal-lonsdale-2020-improving}. While effective in controlled domain-adaptation scenarios, these approaches typically require substantial engineering effort, curated domain resources, or retraining of the base model, and often introduce trade-offs between overall translation fluency and adherence to domain constraints.
DragFT \cite{zheng2024finetuninglargelanguagemodels} targets terminology translation as a core weakness of LLM-based domain-specific MT. The framework integrates dictionary-enhanced prompting, RAG-based few-shot example selection, and parameter-efficient fine-tuning to inject domain terminology directly into the training signal. Its key component, \textit{Dict-rephrasing}, replaces source-side domain terms with their target-language equivalents, allowing LLMs to learn terminology usage in context without increasing prompt length or training data volume. Experiments across IT, law, and medical domains show substantial improvements in word-level translation quality and terminology success rates, with analysis confirming that dictionary-enhanced fine-tuning is more effective than prompting alone, even for strong LLM backbones 
\paragraph{E-commerce and Finance}
Machine translation in terminology-intensive domains such as e-commerce and finance poses persistent and systematic challenges that expose the limits of both conventional NMT systems and general-purpose LLMs. E-commerce text is characterized by stacked keyword phrases, non-canonical title grammar, rapid product turnover, and a high incidence of domain-specific and OOV terms, which often cause models to omit, hallucinate, or paraphrase critical product terminology. One line of work proposes a RAG framework for e-commerce product title translation, where bilingual product titles, descriptions, and bullet points are retrieved from large catalogs and injected as few-shot exemplars into LLM prompts \cite{zhang2024enhancingecommerceproducttitle}. This strategy grounds generation in domain-specific terminology, preserves brand names, and enforces catalog-style conventions without parameter updates, yielding chrF gains of up to 15.3\%, especially for language pairs poorly covered by the base LLM. In contrast, G2ST \cite{10.1145/3589335.3651510} demonstrate that reliable e-commerce MT requires parameter-level specialization through a two-stage fine-tuning process that first injects domain terminology via vocabulary expansion and bilingual term supervision, then adapts models to e-commerce-specific syntax using real titles, with a self-contrastive R-Drop objective to stabilize predictions under dense keyword repetition. Their results consistently show that domain-specialized LLMs outperform prompted ones, highlighting the limitations of zero-shot and few-shot approaches. Complementing this, \citet{GAO2024125087} provide a focused empirical analysis of terminology learning in e-commerce MT by introducing a 20k Chinese–English term lexicon and a 7k parallel title corpus capturing idiomatic expressions, brand names, and morphological variants, and show that tokenizer expansion combined with two-stage full-parameter fine-tuning and self-contrastive regularization yields substantial gains under heavy terminology stacking. Similar limitations surface even more starkly in financial MT, where translation errors carry higher stakes due to dense technical jargon, numerically grounded statements, culturally specific instruments, and a low tolerance for semantic drift or hallucination. As shown by \citet{oncevay-etal-2025-impact}, who introduce the multilingual FinBENCH benchmark with a dedicated financial translation component, state-of-the-art LLMs struggle to preserve financial terminology and numerical fidelity, lack robustness to subtle contextual shifts, and frequently produce overly generic or hallucinated outputs, with degradation particularly severe in lower-resource language directions. Taken together, findings across e-commerce and finance converge on a consistent conclusion: despite strong general translation capabilities, LLMs require explicit domain grounding, either through retrieval-augmented prompting or deeper terminology-aware model specialization, along with domain-specific evaluation, to achieve reliable MT performance in terminology-dense and high-stakes settings.
\paragraph{Translation of Clinical Text}
Accurate translation of clinical text is critical for patient safety, informed decision-making, and equitable access to healthcare, as even minor lexical or factual errors can lead to misinterpretation of diagnoses, treatments, or instructions with potentially severe consequences. Recent work consistently shows that general-purpose MT and LLM-based systems remain fragile in this high-stakes setting. \citet{pakull-etal-2025-preliminary} evaluate an open-source LLM for lay translation of German clinical documents and find that zero-shot and few-shot prompting substantially improve readability and perceived comprehensibility, but still suffer from omissions, residual technical terminology, and limitations in factual correctness, underscoring the need for clinician oversight. Focusing on low-resourced languages, \citet{nigatu-etal-2025-viability} provide a clinically grounded taxonomy of MT errors and demonstrate that mistranslation of medical terminology and omission errors are the primary drivers of high and life-threatening clinical risk, with pre-translation interventions such as paraphrasing or pivoting proving insufficient when the underlying MT system lacks domain-specific vocabulary. Complementing these findings, \citet{CHEN2025151} show that terminology errors in clinical translation are systematic rather than sporadic, with consistent mistranslations disproportionately affecting patient–physician communication and amplifying downstream clinical risk. Collectively, these studies highlight that improving surface-level fluency or accessibility is insufficient for clinical MT, and that reliable deployment requires explicit handling of medical terminology, domain-adapted modeling, and evaluation frameworks centered on clinical risk rather than generic translation quality.
\paragraph{LLMs vs.\ Traditional Models.}
LLMs substantially alter the terminology translation landscape by enabling inference-time constraint incorporation via prompting, in-context learning, or glossary-based instructions, without architectural modification or retraining. Empirical evidence from the WMT25 terminology shared task \cite{semenov-etal-2025-findings} shows that LLM-based systems achieve near-perfect terminology accuracy (often exceeding 97\%) in sentence-level translation tasks when provided with small terminology dictionaries. This result effectively saturates sentence-level benchmarks, indicating that terminology adherence is no longer a bottleneck for modern LLMs in short-context settings.

However, document-level terminology translation remains challenging. In the WMT25 document-level finance track, terminology accuracy for LLM-based systems drops to the 70--80\% range, revealing persistent difficulties in maintaining correct and consistent term usage across long documents with large, one-to-many terminology dictionaries. While LLMs outperform traditional NMT systems in this regime, the substantial gap between sentence- and document-level performance highlights the need for explicit long-context terminology tracking and control mechanisms \citep{oncevay-etal-2025-impact,oncevay-etal-2025-translating}. Notably, high-performing LLM-based systems often improve both overall translation quality (chrF++) and terminology accuracy simultaneously, with no clear negative correlation, contradicting earlier assumptions about inherent quality--terminology trade-offs.

\paragraph{Summary.}
Overall, LLMs decisively outperform traditional MT models in terminology translation at the sentence level and under inference-time constraints. Nevertheless, terminology translation in realistic, document-level, and domain-intensive settings such as finance and e-commerce remains unsolved for both paradigms, positioning long-context terminology control as a central open problem in modern machine translation research.

\subsection{Performance on Low-Resource Languages}
Across surveyed evaluations, the effectiveness of LLM-based MT methods for low-resource languages is highly uneven. Strong gains are primarily observed for languages that are typologically or lexically related to high-resource languages present in LLM pretraining data, share scripts or morphological patterns, or benefit from translation into English \cite{zhu-etal-2024-multilingual,zebaze-etal-2025-context}. In contrast, performance degrades substantially for structurally distant or extremely low-resource languages. These findings indicate that current LLM-based approaches do not uniformly overcome data scarcity, but instead leverage existing cross-lingual support, with data regime and typological proximity playing a more decisive role than prompting or reasoning strategies.

\subsection{LLMs for attribute-controlled MT}
Recent advances in LLMs have given rise to an instruction-controlled paradigm in machine translation, where translation is no longer treated as a fixed mapping between source and target languages, but as a conditional generation task guided by user-specified constraints. In contrast to earlier MT systems where properties such as formality, gender realization, or domain specificity required dedicated model architectures, control tokens, or carefully curated parallel datasets LLMs enable such control to be expressed directly in natural-language prompts. 

A central instance of instruction-controlled MT is gender-controlled translation. GENDEROUS dataset \cite{hackenbuchner-etal-2025-genderous} was introduced in 2025 containing gender-ambiguous sentences. The authors in \cite{sanchez-etal-2024-gender} show that LLMs can produce gender-specific translations by conditioning on few-shot examples, effectively learning \textit{translate using masculine/feminine forms} at inference time.
In Gender-of-Entity (GoE) \cite{huang2025attribute} prompting, LLMs are explicitly instructed to translate a sentence using specified genders for particular entities, provided directly in the prompt (e.g., \textit{for [ENT\_i], use [GENDER\_i]}). Importantly, they also demonstrate that prompt granularity matters: specifying only ambiguous entities can interfere with the realization of other entities' gender. 

A second major axis of instruction-controlled MT concerns style and register, particularly formality. While prior work on stylized MT relied on formality tags or style-labeled corpora, recent approaches leverage LLMs’ instruction-following abilities to impose style constraints. The INMTF framework explicitly uses LLMs as (i) revisers that post-edit NMT outputs to enforce formality, and (ii) reward models that score formality during reinforcement learning, improving both style accuracy and translation quality \cite{cmc.2024.058248}.

These strands converge in recent work on preference-aligned machine translation, which generalizes formality and tone into broader user-specific preferences. PMMT \cite{sun2024pmmtpreferencealignmentmultilingual} explicitly observes that users may prefer different lexical or stylistic realizations (e.g., more polite expressions) that conflict with default semantic mappings and are difficult to encode in traditional MT systems. To address this, PMMT uses LLMs to generate diverse translation candidates and trains a reward model to select outputs aligned with specific preferences, producing large-scale preference-conditioned corpora that can be distilled into smaller MT models for deployment. 
Taken together, these works are not isolated contributions but collectively define an emerging paradigm in LLM-based MT: translation as instruction-following generation. Gender realization, ambiguity handling, formality, and user preference are all expressed as natural-language constraints at inference time, shifting customization from model architecture and data design to prompt specification.

\begin{table}[t]
\centering
\small
\resizebox{\textwidth}{!}{%
\begin{tabular}{p{3.5cm}p{4cm}p{3.2cm}p{2cm}p{3cm}p{2cm}}
\toprule
\textbf{System / Model} &
\textbf{MT-Specific Training} &
\textbf{Language Coverage} &
\textbf{Model size} &
\textbf{Key MT Characteristics} &
\textbf{Openness} \\
\midrule

ALMA / ALMA-R / X-ALMA \cite{xu2024a}  &
Monolingual adaptation + large-scale parallel MT fine-tuning; preference optimization &
50 languages & 7B,13B &
First systematic MT tuning recipe for LLMs; strong MT at small scales &
Open \\
Lingual-SMoE \cite{zhao2024sparse} &
Sparse MoE with language-aware routing and dynamic expert allocation &
100 &
Dense + sparse experts &
Language-level then token-level routing; improved low-resource balance &
Open \\

Hunyuan-MT \cite{zheng2025hunyuanmttechnicalreport} &
Continued pretraining + supervised MT fine-tuning + RL; Chimera ensemble &
33 languages & 1.8B,7B &
Strong multilingual and low-resource MT &
Open \\

Qwen-MT  \cite{yang2025qwen3technicalreport}&
Large-scale parallel MT fine-tuning + reinforcement learning &
92 languages & NA & 
Industrial MT LLM with terminology control and domain prompts &
API / Partial \\

Tower / Tower+ \cite{alves2024tower,rei-etal-2024-tower,rei2025towerbridginggeneralitytranslation}  &
Continued multilingual pretraining + MT instruction fine-tuning &
15 languages & 2B,7B,9B,
13B,72B &
Translation-centric instruction tuning &
Open \\

Seamless M4T v2 \cite{communication2023seamlessmultilingualexpressivestreaming} & Multimodal training including speech and text data & 200 languages & 1.2B,2.3B & Unified multilingual MT trained jointly with speech, giving robust low-resource and zero-shot translation & Open \\
GemmaX2 \cite{cui-etal-2025-multilingual} &
Parallel-first continued pretraining + MT fine-tuning &
28 languages & 2B,9B &
Open LLMs competitive with large proprietary MT systems &
Open \\

MT-R1-Zero \cite{feng-etal-2025-mt-r1}   &
MT-specific tuning with rule-based rewards &
NA & 7B & 
RL-based MT specialization without human labels &
Open \\

SSR-Zero \cite{yang2025ssrzerosimpleselfrewardingreinforcement}  &
MT-specific tuning with self-reward &
NA & 7B &
RL-based MT specialization without human labels &
Open \\

\midrule
\multicolumn{6}{c}{\textbf{General-purpose LLMs with Strong Performance in Machine Translation}} \\
\midrule

Aya \cite{ustun-etal-2024-aya} &
Massively multilingual instruction tuning with human-curated parallel and preference data &
101 &
13B &
Strong zero-shot and low-resource MT; explicitly optimized for multilingual alignment &
Open \\

ChatGPT \cite{openai2024gpt4technicalreport} &
Instruction tuning with RLHF on large-scale multilingual data &
100+ &
Undisclosed &
High fluency and adequacy; competitive with commercial MT in high-resource settings &
API \\

Gemini \cite{geminiteam2025geminifamilyhighlycapable,comanici2025gemini25pushingfrontier} &
Multimodal and multilingual instruction tuning with long-context alignment &
100+ &
Undisclosed &
Strong document-level MT and long-context consistency; robust multilingual transfer &
API \\

Claude \cite{TheC3} &
Constitutional AI with multilingual instruction tuning &
50+ &
Undisclosed &
High discourse-level fluency and coherence; limited coverage for very low-resource languages &
API \\

\bottomrule
\end{tabular}%
}
\caption{Major LLM-based MT systems; Note that we report a very few selected models; we also exclude the papers which does not have either of API access or open weights.}
\label{tab:mt_systems_corrected}
\end{table}

\section{LLMs and Encoder-Decoder across Languages}
This section explores how LLMs perform across languages and how they compare to dedicated MT systems. An early study found that ChatGPT lags behind traditional MT systems such as Google Translate, DeepL, and Tencent, but observed that GPT-4 performs competitively with these systems in zero-shot settings \cite{jiao2023chatgptgoodtranslatoryes}. This indicates that some LLMs achieve performance on par or even better with traditional MT systems for high-resource languages. 

When translating from Spanish to 11 indigenous South American languages, fine-tuned M2M100 was shown to consistently outperform GPT-4 \cite{stap-araabi-2023-chatgpt}. The performance gap further increases when M2M100 is combined with techniques like kNN-MT \cite{khandelwal2021nearest}. The contrast between high-resource and low-resource languages is also highlighted by a multitask evaluation, where ChatGPT's performance, although slightly lower, remains close to fine-tuned state-of-the-art models \cite{bang-etal-2023-multitask}. For instance, ChatGPT achieves 58.64 chrF++ compared to 63.5 for fine-tuned models in XX$\rightarrow$En translation, and 51.12 chrF++ versus 54.4 for En$\rightarrow$XX \cite{nllbteam2022languageleftbehindscaling}. However, for low-resource languages, the difference becomes much more pronounced, with ChatGPT scoring 27.79 chrF++ compared to 54.9 for fine-tuned models in XX$\rightarrow$En, and 21.57 versus 41.9 chrF++ for En$\rightarrow$XX \cite{popovic-2015-chrf}. A similar study \cite{song2025llmsilverbulletlowresource} identifies key bottlenecks, such as the asymmetry in translation quality between high-resource to low-resource directions and the poor generative capability of smaller models. The study finds that LLMs generally underperform for low-resource languages compared to traditional encoder-decoder NMT systems, which is also confirmed by \cite{robinson-etal-2023-chatgpt} earlier. Further, they observe that generating En $\rightarrow$ XX is more difficult compare to generating XX$\rightarrow$En.

Recent work has explored whether LLMs can perform translation in extreme low-resource settings where parallel data is virtually absent. The MTOB benchmark \cite{tanzer2024a} evaluates this scenario by requiring models to learn translation for a previously unseen language using only a single grammar book, a bilingual lexicon, and a very small parallel corpus. While long-context in-context learning substantially improves LLM performance, even strong models with large context windows remain well below a data-matched human baseline. The study shows that translation improvements mainly stem from parallel examples and lexical resources, whereas grammatical explanations alone contribute little to translation quality. Overall, these findings indicate that encoder–decoder MT models with explicit parallel supervision remain more effective than LLM-based approaches in extremely low-resource translation settings.
Similarly, the authors in \cite{aycock2025can} split grammar books into parallel examples and grammatical explanations, then tested LLMs and a fine-tuned NMT model on translation, grammatical judgment, and gloss prediction tasks. They also introduced typological prompts summarizing high-level grammatical features.
Their findings show that translation improvements come almost entirely from parallel examples, not grammatical explanations. Fine-tuned models performed as well or better than LLMs using grammar books. However, for linguistic tasks like grammaticality judgment and gloss prediction, typological prompts combined with parallel examples led to strong performance. The study concludes that for translation, collecting parallel data is more effective than relying on descriptive grammar, while structured grammatical knowledge is better suited for non-translation tasks.
The authors in \cite{zheng2025asymmetricconflictsynergyposttraining} investigates the persistent issue of the Curse of Multilinguality (CoM) in LLM-based multilingual MT, where scaling to more languages introduces performance bottlenecks. The authors identify a key phenomenon: asymmetry in linguistic conflicts and synergies: with translation from other languages to English (XX$\rightarrow$En) suffering from conflicts, while English to other languages (En$\rightarrow$XX) benefits from synergy. These issues primarily arise during the post-training phase rather than during multilingual pretraining. To address this, they propose a Direction-Aware Training (DAT) approach: applying separate training for XX$\rightarrow$En directions to reduce conflicts and group multilingual training for En$\rightarrow$XX to leverage synergy. They further introduce group-wise model merging (DATM) to improve scalability by reducing the number of LoRA adapters while maintaining translation quality. This work highlights the importance of targeted post-training strategies and provides a path toward building high-quality, cost-effective MT systems.
\begin{tcolorbox}[colback=yellow!20,colframe=black!75]
LLMs demonstrate competitive performance with traditional MT systems in high-resource and zero-shot scenarios, with GPT-4 approaching or matching systems like Google Translate and DeepL. However, in low-resource and extremely low-resource languages, encoder-decoder models like fine-tuned M2M100 consistently outperform LLMs, especially when enhanced with techniques like kNN-MT. Translation quality also varies asymmetrically by direction: XX$\rightarrow$En often suffers more than En$\rightarrow$XX due to inherent linguistic conflicts.
The Curse of Multilinguality persists in LLM-based MT, particularly in XX$\rightarrow$En directions, due to post-training conflicts. This has led to innovations like Direction-Aware Training (DAT) and group-wise model merging (DATM) to manage conflict-synergy dynamics and improve scalability with fewer parameters.
\end{tcolorbox}
Towards building efficient LLM-based MT systems, a pre-trained decoder-only LLM can serve as the encoder, paired with a lightweight NMT-style decoder \cite{luo2025decoderonlylargelanguagemodels}. To enable compatibility between the LLM's output and the decoder's input, the intermediate representations from multiple LLM layers are integrated, transformed to a lower dimensional space, and optionally enhanced with additional encoder layers to capture bidirectional context. Training proceeds in two phases: initially, only the transformation layers and decoder are trained while keeping the LLM frozen; subsequently, the entire model is fine-tuned on ComMT, a comprehensive benchmark covering general, domain-specific, document-level, terminology-constrained, and post-editing translation tasks. This approach, called LaMaTE, demonstrates strong performance and generalization, while significantly improving efficiency achieving up to 6.5× faster decoding and 75\% reduction in memory usage, highlighting the potential of LLMs as encoders in NMT architectures.

A key distinction between LLMs and encoder-decoder models is their ability to generalize beyond sentence-level translation, offering rich contextualization, stylistic control, and flexibility via prompting. A recent study \cite{stap-etal-2024-fine} reveals that this distinction erodes when LLMs are fine-tuned on parallel data. In many respects, fine-tuned LLMs begin to resemble traditional NMT systems: they lose control over output formality, struggle with context-sensitive disambiguation, and fail to generalize effectively to unseen domains without explicit retraining. While idiomatic translation often improves when initial performance is weak, other capabilities such as formality control and document-level contextualization, suffer significant degradation.
This degradation is not readily apparent when relying solely on standard evaluation metrics. For instance, COMET scores may remain stable or even improve while the model's ability to follow stylistic constraints or disambiguate based on discourse context declines. This indicates a mismatch between what current MT metrics measure and what users value in real-world translation tasks. This convergence suggests that careless adaptation of LLMs risks negating their primary benefits. While LLMs initially surpass NMT systems in their versatility, indiscriminate fine-tuning can narrow their capabilities to those of more brittle translation architectures. This finding reinforces the need to develop fine-tuning strategies that preserve the emergent capabilities of LLMs rather than inadvertently discarding them.

\section{LLM-Based Evaluation of Machine Translation}

LLMs are increasingly being used as automatic judges of MT quality. Unlike traditional metrics like BLEU or learned metrics like COMET and BLEURT, LLM-based evaluators rely on prompting a generative model with source, translation, and possibly reference inputs to elicit quality scores or error analyses. While this LLM-as-judge paradigm offers flexibility and strong system-level performance, it also raises critical concerns regarding bias, segment-level reliability, and sensitivity to prompt design. This section surveys the state of the art in LLM-based MT evaluation, synthesizing benchmark results, methodological advances, and known limitations.
\begin{tcolorbox}[colback=yellow!20,colframe=black!75]
Recent studies show that LLMs can act as effective MT evaluators, achieving strong correlation with human judgments at the system level, especially when guided by carefully designed prompts. Error-focused prompting and MQM-style span annotation substantially improve segment-level reliability and provide interpretable feedback beyond scalar scores. However, LLM-based evaluators exhibit systematic biases, including sensitivity to prompt phrasing, input order, and verbosity, which directly affect score stability. Structured approaches such as multi-agent debate and contrastive scoring consistently outperform single-pass LLM judging and reduce these biases. Likelihood-based and preference-oriented LLM metrics capture complementary quality dimensions that differ from lexical and neural metrics. Overall, the key finding is that LLM-based evaluation is most effective when explicitly structured and constrained, and should be used alongside, not instead of, established MT metrics. We refer the readers to \cite{10650457} for a focused study on evaluation.
\end{tcolorbox}
\subsection{Performance Benchmarks and Correlation with Human Judgments}

Initial studies demonstrated that LLMs can outperform traditional MT metrics under carefully designed setups. GEMBA-style prompting~\cite{kocmi-federmann-2023-gemba, kocmi-federmann-2023-large} is a template-based approach to generate adequacy and fluency scores from GPT-3.5 and GPT-4. Results from the WMT22 Metrics task show that GPT-4 achieves the highest system-level pairwise accuracy and Kendall’s-$\tau$ for English--German and English--Russian. However, COMET outperforms GPT-4 on Chinese--English, suggesting that performance depends on both language pair and evaluation level. These findings align with broader studies showing that LLM evaluators often perform well in system-level aggregation but struggle with sentence-level reliability~\cite{fernandes-etal-2023-devil}.
To address segment-level limitations, Lu et al. \cite{lu-etal-2024-error} proposed \textit{Error Analysis Prompting}, which combines MQM-style error detection CoT reasoning. This method improves segment-level alignment with human scores and helps bridge the gap with learned metrics like COMET, BLEURT, and MetricX. Similarly, AutoMQM \cite{fernandes-etal-2023-devil} explicitly prompts LLMs to identify error spans, categories, and severities. These annotations are then converted into MQM scores algorithmically. The approach achieves state-of-the-art segment-level correlation in zero-shot setups and offers interpretable feedback. The authors also show that model scale matters: larger LLMs benefit more from error-centric prompting, while smaller ones require fine-tuning.
Building on this line of research, GPTScore ~\cite{fu-etal-2024-gptscore} proposes a likelihood-based evaluation metric derived from instruction-tuned LLMs, which computes the conditional generation probability of a translation given the source or reference. On the MQM-2020 dataset \cite{mathur-etal-2020-results}, GPTScore achieves higher Spearman correlation with human scores than conventional metrics such as ROUGE, BERTScore, and BARTScore. However, this comparison excluded more recent learned baselines like COMET and focused primarily on high-resource language pairs (e.g., English--Chinese), limiting its generalizability.

Reference-based metrics introduces challenges: high-quality references are expensive to obtain, and valid translations that diverge lexically from the reference may be unfairly penalized. Moreover, reliance on a single reference can bias evaluators toward literal or overly conservative translations. However, reference based methods have higher correlation with human judgements in traditional metrics.
Similarly, Qian et al. \cite{qian-etal-2024-large} explored what kinds of input information LLMs need to evaluate translations accurately. They found that including both the source and reference translation is crucial, whereas detailed annotation guidelines or error examples have diminishing returns. Further, they noted that CoT prompting improves performance, however LLMs still show inconsistency in producing stable numerical scores, posing challenges for deployment in automated pipelines.

The WMT25 Translation Quality Evaluation Shared Task \cite{lavie-etal-2025-findings} found that large LLM-as-a-judge systems achieve strong performance at the system level, whereas reference-based baseline metrics outperform LLMs at the segment level when correlating with human judgments, suggesting that different error phenomena are captured at different evaluation granularities. 
The authors in \cite{10.1609/aaai.v39i22.34522} show that corpus-level aggregation can substantially mask sentence-level variance due to the classical ratio-of-averages versus average-of-ratios effect, leading to inflated system-level stability even when individual segment scores are noisy or inconsistent. Although this result is demonstrated for lexical metrics, the same aggregation mechanism is especially consequential for LLM-based evaluators. Unlike lexical or learned metrics, LLM judges produce sentence-level scores that is sensitive to local phrasing, paraphrastic choices, and reference divergence. When aggregated over a test set, these sentence-level inconsistencies are smoothed out, while broader distinctions between stronger and weaker MT systems are preserved. This helps explain why LLM-based evaluators achieve high correlation with human judgments at the system level, yet fail to exhibit corresponding reliability at the segment level, where individual judgment inconsistencies are no longer mitigated by aggregation.
Recent empirical analyses further demonstrate that different automatic evaluation metrics can encode fundamentally different notions of translation quality, leading to systematic disagreements that persist even at scale. The authors in \cite{gisserot-boukhlef-etal-2024-preference} show that neural metrics such as xCOMET-QE and CometKiwi exhibit strong internal consistency but can diverge sharply from lexical metrics like chrF, particularly for out-of-English translation directions. Their results indicate that improvements measured by neural metrics do not necessarily translate into gains in surface-level adequacy or lexical faithfulness, highlighting that metric agreement cannot be assumed even when evaluating the same model outputs. These findings reinforce the view that neural and LLM-driven evaluators capture complementary but incomplete dimensions of translation quality, and that relying on a single metric may obscure important trade-offs. Consequently, robust evaluation of LLM-based MT systems requires multi-metric reporting and, where possible, validation against human judgments rather than sole dependence on neural or LLM-based evaluators.

\subsection{Biases in LLM-as-Judge Setups}

Despite their promising performance, LLM-based evaluators exhibit several biases that can compromise their reliability. Although there have been very limited study on biases for MT in LLM-as-Judge setup, we discuss a few major studies that are applied in similar NLP tasks and could be relevant.

\textbf{Prompt Sensitivity and Input Order Bias.} 
LLM outputs vary significantly with prompt phrasing. Studies show that even the order of presenting system outputs in a pairwise comparison can skew evaluations \cite{ye2024justiceprejudicequantifyingbiases,shi2025judgingjudgessystematicstudy}. 

\textbf{Verbosity Bias.} It has been observed that LLMs often assign higher scores to longer translations, mistaking verbosity for completeness. This bias not only inflates scores for verbose outputs but also undermines fair comparison against concise, high-quality translations \cite{ye2024justiceprejudicequantifyingbiases, briakou2024implicationsverbosellmoutputs}. 

These biases underscore the importance of treating LLM-as-judge evaluations as protocol-sensitive rather than model-intrinsic. Reliable use therefore requires explicit safeguards, including controlled prompt templates, randomized ordering of candidate translations, and evaluation formats that discourage reliance on superficial cues such as length. In practice, structured or error-focused prompting and aggregation over multiple judgments can further reduce sensitivity to individual prompt realizations. Without such precautions, reported gains from LLM-based evaluators may reflect artifacts of the evaluation setup rather than genuine improvements in translation quality.
\subsection{Structured and Multi-Agent Evaluation Frameworks}

To mitigate the above biases and enhance reliability, recent work has proposed structured multi-agent or contrastive frameworks:

\textbf{M-MAD (Multidimensional Multi-Agent Debate).} Feng et al. \cite{feng2025mmadmultidimensionalmultiagentdebate} introduce a three-stage evaluation pipeline where separate LLM agents assess MQM dimensions such as fluency, adequacy, style, and terminology. Each dimension is scored via a two-agent debate, followed by aggregation by a judge agent. This framework achieves higher system-level correlation than previous single-agent approaches and improves segment-level stability, though it still trails top learned metrics in fine-grained assessment.

\textbf{Contrastive Metrics} ContrastScore \cite{wang2025contrastscorehigherqualitybiased} propose a contrastive scoring method that compares the outputs of strong and weak LLMs. By examining probability differences, the approach discourages overly generic or verbose responses and promotes high-quality, nuanced outputs. Despite using relatively small models (e.g., 3B and 0.5B), ContrastScore matches or exceeds the performance of larger single-model evaluators while reducing evaluation cost.

\subsection{Summary}
Importantly, segment-level unreliability of LLM-based MT evaluators should not be interpreted as frequent hallucinations or fluency degradation. Empirical analyses consistently show that strong LLM translations are fluent and largely faithful, yet exhibit legitimate sentence-level variation in lexical choice, paraphrasing, explicitation, and stylistic realization. These variations often diverge from reference translations without constituting genuine errors, but are nevertheless penalized by both reference-based metrics and LLM-as-judge systems. Furthermore, LLM evaluators show sensitivity to prompt phrasing, verbosity, and local surface cues, leading to unstable scalar scores for near-equivalent outputs. While these effects introduce high variance at the segment level, they are largely smoothed out through corpus-level aggregation, resulting in strong system-level correlation with human judgments. 
LLM-based evaluators, while sometime competitive to traditional metrics, they still suffer from several limitations:

\begin{itemize}
    \item \textbf{Segment-Level Fragility:} Most LLM metrics struggle to provide stable segment-level scores without elaborate prompting or multi-agent feedback.
    \item \textbf{Computational Cost:} Querying large LLMs for every translation segment is expensive, especially for large-scale evaluations.
    \item \textbf{Prompt Sensitivity:} Minor changes in prompt phrasing can shift scores, challenging reproducibility.
    \item \textbf{Biases:} Fluency and verbosity biases persist even in advanced setups.
\end{itemize}

Despite these issues, LLM-based MT evaluators offer advantages in interpretability. Their ability to explain judgments, adjust to task-specific criteria, and handle reference-free setups make them a valuable complement to learned metrics. Moreover, hybrid methods that combine LLM reasoning with learned metric supervision are a promising direction.

\section{LLMs Beyond Sentence-level Translation}
\subsection{Document Translation}
Early studies on document-level machine translation (DocMT) with LLMs suggested that LLMs could substantially advance discourse-aware translation. DocMT using GPT-3.5 and GPT-4 outperform strong commercial translation systems and advanced document-level neural machine translation approaches in human evaluation across multiple domains and language pairs \cite{wang-etal-2023-document-level}. However, they also observe that gains from explicitly context-aware prompting are modest and highly sensitive to prompt design (also observed in \cite{wu-hu-2023-exploring}), suggesting that strong DocMT performance does not necessarily imply systematic exploitation of document context.
Subsequent work investigates whether document-level behavior can be more reliably induced through supervised adaptation. The authors in \cite{wu2024adaptinglargelanguagemodels} show that moderately sized LLMs, when fine-tuned on document-level parallel data, can match or even surpass much larger proprietary models on several DocMT benchmarks, particularly for coherence and discourse-sensitive phenomena. At the same time, the study identifies a critical limitation: autoregressive error propagation when previously generated translations are reused as context, which can lead to off-target translations.
A controlled perturbation and attribution analysis and find that most evaluated LLMs are largely insensitive to document context, even for phenomena such as pronoun resolution that require cross-sentence information \cite{mohammed-niculae-2025-context}. Translation quality remains stable when gold context is replaced with perturbed or random context, and attribution methods show that relevant antecedents receive minimal contribution during generation. These results indicate that strong DocMT scores alone are insufficient evidence of effective context usage. From a complementary evaluation perspective, the author in \cite{kim-2025-context} revisits document-level assessment and shows that while contextual information is theoretically present in all sentences, it rarely changes human judgments in practice. Through a detailed human study, it demonstrates that only a limited portion of overall translation quality variance can be attributed to document-level factors, challenging assumptions that DocMT performance is dominated by context-dependent phenomena.
\begin{tcolorbox}[colback=yellow!20,colframe=black!75]
Early work shows that LLMs can achieve strong document-level MT performance, often outperforming commercial and neural DocMT systems, but high scores do not necessarily imply effective use of document context. Empirical analyses reveal that LLMs are frequently insensitive to cross-sentence information, and gains from naïve long-context prompting are modest and prompt-dependent. More reliable document-level behavior emerges when context is explicitly structured, selectively retrieved, or incrementally tracked. Training-time interventions typically involve mixing sentence- and document-level instructions or limited document-parallel data, not drastic changes in objectives or architectures.
\end{tcolorbox}
A separate line of work focuses on identifying structural reasons for weak document-level behavior in LLM-based translation systems. Li et al. attribute the failure of lightweight LLMs in DocMT to their reliance on sentence-level translation instructions during fine-tuning \cite{LI2026132041}. They show that translation quality degrades sharply beyond 512 tokens when models are trained only on sentence-level instructions, but can be stabilized up to 2048 tokens by mixing sentence- and document-level translation instructions. Importantly, they find that overemphasizing sentence-level instructions can actively harm document-level translation, highlighting a tension between sentence-level accuracy and document-level coherence. Their results suggest that DocMT capability does not emerge automatically from scale, but must be explicitly induced during training.

Other approaches attempt to enforce document-level context by integrating a document-level language model with a sentence-level translation system, while explicitly neutralizing the translation model’s internal language model to avoid overcounting source-independent probabilities \cite{petrick-etal-2023-document}. Their experiments show consistent improvements on document-targeted metrics and discourse-sensitive phenomena without requiring document-level parallel training data, although they note that back-translation remains a stronger but more computationally expensive alternative.
Recent work argues that indiscriminately providing long document context is suboptimal, as useful discourse dependencies are sparse. The authors in \cite{peng-etal-2025-self} propose a self-retrieval framework for DocMT that dynamically selects relevant local and distant context sentences based on relevance scoring. Using pointwise mutual information for retrieval, they show improvements in both reference-based metrics and lexical consistency, particularly for long documents such as scientific articles. Their results provide direct evidence that document context improves translation quality when it is selectively retrieved rather than uniformly included. Along similar lines, authors in \cite{liu-etal-2025-improving-llm} argue that raw document context should be transformed into structured signals before being used by LLMs. By incorporating document-level summaries and entity translation knowledge, and then fusing multiple knowledge sources through ranking, they achieve consistent gains across multiple models and translation directions, while observing that individual knowledge sources can help some sentences but harm others.
DelTA \cite{wang2025delta} proposes an online formulation of document-level machine translation in which translation is performed incrementally, where each newly translated sentence updates an internal document state that influences future decisions. The system explicitly maintains discourse-relevant information such as lexical choices and coherence cues and allows later translations to adapt based on accumulated context. Empirical results show that this incremental, state-based approach improves document-level consistency and coherence compared to sentence-level translation and naïve sliding-window baselines, particularly for long documents where early lexical or stylistic choices strongly affect downstream quality. The key contribution of DelTA is to demonstrate that document context need not be modeled as a static input sequence; instead, it can be represented as an evolving decision state, making DocMT applicable to streaming and real-time translation scenarios while still preserving discourse-level gains.

Refinement-based approaches \cite{dong2025intermediatetranslationsbetterone} suggest reframing DocMT as a post-editing problem rather than direct generation by refining document translations using two complementary intermediate outputs: sentence-level and document-level translations. They show that sentence-level translations offer better coverage but weaker coherence, while document-level translations improve discourse coherence but suffer from under-translation. Refining both jointly allows the model to leverage their complementary strengths, and a quality-aware weighting strategy further improves performance by focusing learning on difficult cases. These results indicate that explicit comparison and refinement can teach LLMs when and how document context should influence translation decisions.
While refinement-based approaches show that LLMs can effectively exploit document context when comparing alternative translations, they still rely on document-level bilingual data. The authors in \cite{li-etal-2025-enhancing-large} extend this paradigm by demonstrating that document-level post-editing can be learned using only monolingual data, significantly broadening the applicability of refinement-based DocMT.
The Multi-Agent Translation Team (MATT) uses multiple specialized agents such as a translator, evaluator, proofreader, editor, and editor-in-chief that iteratively refine translations until a predefined quality threshold is met \cite{Peter2024MATT}. Experiments on low-resource language pairs show that MATT is consistently preferred by human evaluators over single-agent baselines and, in some cases, over Google Translate, even when automatic metrics favor more literal outputs. Beyond translation quality, the paper highlights a systematic mismatch between human judgments and automated metrics in multi-agent DocMT settings, underscoring the importance of human-centered evaluation. 
It has been found that LLMs can be more reliably used as document-level evaluators rather than generators, at the system level  \cite{kudo-etal-2024-document}. By reranking sentence-level translation candidates sequentially using document context, they achieve improved cross-sentence consistency, indicating that LLMs can exploit document information effectively when restricted to selection rather than generation.

From a systems and training perspective, most LLM-based document-level MT approaches do not introduce fundamentally new data regimes or optimization objectives compared to sentence-level MT. Instead, they extend sentence-level translation pipelines with additional context signals, either at inference time or through limited training-time adaptations such as mixing sentence- and document-level instructions. As a result, improvements in DocMT are often attributable to how document context is selected, represented, and consumed, rather than to large-scale architectural or algorithmic changes.
While we aim to provide a high-level overview of recent advances in document-level and context-aware machine translation, this is a broad area that requires a dedicated survey. We therefore, refer the  readers to \cite{appicharla2025sentencesurveycontextawaremachine}for a more focused and detailed survey. 
\subsection{Literary Translation}
 
Literary translation poses challenges that extend beyond sentence-level adequacy, requiring systems to model long-range discourse, stylistic consistency, and culturally grounded meaning. A central theme across submissions to the WMT 2023 Discourse-Level Literary Translation shared task is the trade-off between \emph{explicit document modeling} and \emph{implicit exploitation of context}, alongside the limitations of automatic metrics for capturing literary quality \cite{wang-etal-2023-findings}.

Several systems investigate document-level modeling within neural machine translation architectures by contrasting sentence-level and paragraph- or document-level training. The MAX-ISI submission shows that aggregating sentences into paragraph-level units improves discourse-sensitive metrics such as d-BLEU and BlonDe \cite{jiang-etal-2022-blonde}, although gains in sentence-level BLEU remain limited \cite{an-etal-2023-max}. Evaluations of the MEGA architecture further suggest that architectural mechanisms for long-range dependency modeling are constrained by the effective context available in training data. Similar observations arise in the MAKE-NMT-Viz system, where fine-tuning mBART50 substantially reduces hallucinations and improves fluency, but concatenation-based document-level modeling yields inconsistent quantitative gains \cite{lopez-etal-2023-make}. Their qualitative analysis highlights persistent challenges in pronoun resolution, tense consistency, and stylistic adequacy, reinforcing the gap between automatic metrics and human judgments.

\begin{tcolorbox}[colback=yellow!20,colframe=black!75]
Research on literary translation is shifting away from sentence-level, architecture-centric NMT toward context-aware, LLM-driven pipelines that operate on paragraphs, documents, or entire books. Gains increasingly come from data quality, long-context prompting, multi-stage generation, and explicit consistency control, rather than from novel model architectures. At the same time, the community recognizes that automatic metrics systematically under-represent literary quality, driving a parallel shift toward human-aligned, reference-free evaluation and preference-based assessment.
\end{tcolorbox}
Other submissions emphasize data-centric strategies over architectural changes. The Huawei HW-TSC system demonstrates that large-scale domain adaptation via forward translation, back-translation, and data diversification delivers the largest improvements for literary translation, while discourse-specific techniques such as multi-resolution document-to-document translation and lexically constrained training primarily improve consistency phenomena such as named entities and zero pronouns \cite{xie-etal-2023-hw}. Model capacity is addressed in the TJUNLP system through a Mixture-of-Experts Transformer initialized from a dense baseline, with results indicating that sparse expert specialization can outperform dense models under limited literary training data \cite{zhu-xiong-2023-tjunlp}.

Beyond conventional NMT systems, multiple papers highlight the effectiveness of large language models for discourse-level literary translation. A large-scale human evaluation shows that GPT-3.5 produces substantially higher-quality translations when translating entire paragraphs rather than isolated sentences across 18 language pairs, reducing mistranslations, grammatical errors, and inconsistencies, though critical errors and omissions persist \cite{karpinska-iyyer-2023-large}. Complementing this analysis, the DUTNLP system demonstrates that carefully designed prompts and discourse segmentation strategies enable GPT-3.5 to outperform fine-tuned document-level NMT baselines without additional training, underscoring prompting as a competitive alternative when resources are constrained \cite{zhao-etal-2023-dutnlp}.

Overall, these studies collectively show that while discourse-level context is essential for literary translation, it remains difficult to exploit reliably. Traditional NMT systems benefit most from domain adaptation with modest gains from document-level modeling, whereas large language models show strong improvements when operating directly on paragraphs. Nonetheless, unresolved issues in stylistic fidelity and subtle discourse phenomena reinforce the continued necessity of human evaluation and fine-grained error analysis.

Complementing system-centric evaluations, recent work has examined literary translation from a user and pedagogical perspective. A study \cite{Abdelhalim31122025} investigates how novice and advanced EFL student translators perceive literary translations generated by ChatGPT and Google Translate. The participants consistently favored ChatGPT for its fluency, stylistic naturalness, and cultural sensitivity, and criticized Google Translate for literalness and failure to convey literary nuance. These findings align with WMT observations that LLM-based systems better capture discourse-level phenomena, while also highlighting that perceived translation quality and usefulness may diverge from automatic metric scores.

Recent work has begun to directly address the inadequacy of existing automatic metrics for literary translation. LITRANSPROQA \cite{zhang-etal-2025-litransproqa}, a reference-free, LLM-based evaluation framework that models the professional assessment process of literary translators through structured question answering. Instead of relying on surface similarity or learned regression alone, LITRANSPROQA evaluates translations along dimensions central to literary quality such as authorial voice, cultural grounding, stylistic coherence, and creative equivalence using a curated set of expert-designed questions. Empirically, LITRANSPROQA has higher correlation with human judgments and markedly improved adequacy in ranking professional human translations above MT outputs. These results underscore that progress in literary translation is tightly coupled with evaluation paradigms that reflect expert literary judgment, reinforcing concerns raised in WMT that conventional metrics systematically undervalue literary nuance.

% Add the benchmark paper here

The WMT 2024 submissions on literary translation \cite{wang-etal-2024-findings} primarily focus on enforcing consistency in entities, terminology, and narrative style across long contexts.
One line of work adopts a modular pipeline centered on lexical and cultural consistency. In CloudSheep system \cite{liu-etal-2024-cloudsheep}, the system focuses on identifying and normalizing proper names and idiomatic expressions prior to translation. Named entities are extracted and mapped to consistent target-language forms using curated glossaries and LLM-assisted validation, while idioms are handled through dictionary grounding followed by contextual paraphrasing. Translation itself is largely sentence-based, with discourse consistency enforced through preprocessing and post-editing rather than joint modeling.
Another approach emphasizes document-level training and inference through chunk-based modeling and multi-model collaboration \cite{sun-etal-2024-final}. Here, the system fine-tunes LLMs using multi-sentence chunks and generates multiple candidate translations using different LLMs. A dedicated editing stage merges these candidates into a single output, followed by an explicit terminology proofreading step that enforces consistent lexical choices across the document.

A complementary direction explores explicit context-aware decoding within a trained model \cite{luo-etal-2024-context}. It combines continued pretraining and SFT with an incremental decoding strategy that conditions each sentence on previously translated context. In addition, stylistic information is retrieved from similar sentences to guide generation, demonstrating that document-level coherence and stylistic consistency can be partially addressed through decoding-time context integration.

Training-free document-level adaptation is investigated in \cite{liu-etal-2024-noveltrans}, where large commercial LLMs are guided using document-level prompting strategies. The system constructs terminology tables to ensure consistent translation of rare and domain-specific expressions and employs multi-aspect prompting to generate diverse translation candidates. A reference-free quality estimation model is then used to select or assemble sentence-level outputs into a coherent document-level translation.

Finally, the authors in \cite{li-etal-2024-linchance} provides insight into practical constraints faced when applying LLMs to literary translation. The study compares fine-tuning and prompting strategies under limited computational resources, showing that parameter-efficient adaptation and careful prompt design can generate competitive performance even when full fine-tuning is infeasible. This work underscores the trade-offs between model capacity, resource availability, and translation quality in real-world literary MT settings.

Recent work further explores whether literary translation quality can be improved by explicitly modeling human collaborative workflows rather than relying on a single model. TRANSAGENTS \cite{10.1162/TACL.a.25} is a multi-agent LLM framework that simulates a professional translation company, incorporating roles such as translators, editors, localization specialists, and proofreaders. The system translates ultra-long literary texts chapter by chapter, guided by shared glossaries, stylistic guidelines, and long-term narrative memory.
Although TRANSAGENTS achieves lower document-level BLEU scores, human and LLM-based preference evaluations consistently favor its outputs over both GPT-4 and professional human references. This discrepancy highlights once again the limitations of surface-level metrics and suggests that multi-stage, role-specialized generation pipelines may better capture literary coherence, stylistic richness, and cultural adaptation particularly for long-form narratives.

Overall, these collectively illustrate that progress in literary translation with LLMs is driven less by novel architectures and more by effective use of long-context prompting, document-level constraints, and multi-stage generation pipelines. They reinforce the view that literary translation serves as a natural benchmark for evaluating discourse-level capabilities of LLM-based MT systems beyond sentence-level translation.

\section{Summary and Recommendation}
This survey reviewed recent advances in LLM based MT and analyzed how these models are reshaping the methodological and conceptual foundations of the field. The evidence shows a clear shift from purely supervised translation pipelines toward instruction-driven generation frameworks that emphasize flexibility, controllability, and rapid adaptation. Techniques such as synthetic data generation, in-context learning, dictionary- and retrieval-augmented prompting, and parameter-efficient fine-tuning have reduced the cost of adapting systems to new languages and domains, while enabling richer forms of control at inference time.

Despite these advances, LLM based MT does not uniformly surpass traditional encoder–decoder systems. Strong and sometimes near-parity performance is consistently observed for high-resource language pairs and zero-shot settings. However, for structurally distant and extremely low-resource languages, encoder–decoder models trained with explicit parallel supervision remain more reliable. In such settings, gains attributed to sophisticated prompting or reasoning strategies are typically marginal compared to improvements obtained from higher-quality bilingual or preference-aligned signal. This highlights that data regime and linguistic coverage remain the dominant factors governing translation quality.

A major trend emerging across recent work is the growing role of preference-based optimization. Objectives based on contrastive preference optimization, reinforcement learning, and self-rewarding schemes provide a more expressive alignment signal than reference likelihood alone. These approaches capture subtle differences in adequacy, fluency, formatting, and human preference that are difficult to encode with supervised fine-tuning. Preference-based methods are particularly valuable in low-resource and high-stakes domains, where carefully constructed synthetic or weakly supervised preference data can partially compensate for the lack of large-scale human annotation.

Evaluation has emerged as a critical bottleneck. LLM based evaluators achieve strong correlation with human judgments at the system level, but remain unstable at the segment level and are sensitive to prompt design, verbosity, and local surface cues. Learned metrics such as COMET continue to provide more consistent fine-grained assessment, while LLM based evaluation is best viewed as complementary, offering interpretability, reference-free analysis, and structured error explanations. As translation quality improves, limitations of current evaluation protocols increasingly constrain measurable progress, especially for discourse-level, stylistic, and literary translation.

Beyond sentence-level translation, LLMs demonstrate clear advantages in document-level and literary translation when combined with long-context prompting, selective context retrieval, and multi-stage or multi-agent workflows. However, strong document-level scores alone do not imply systematic exploitation of discourse context. Empirical analyses show that explicit mechanisms for context selection, tracking, and consistency enforcement are still required. Terminology translation exemplifies this gap: sentence-level terminology control is largely saturated under inference-time constraints, while document-level terminology consistency remains an open challenge for both LLM based and traditional systems.

Based on the surveyed literature, we summarize the main recommendations for future research as follows:
\begin{itemize}
    \item Low-resource MT should prioritize signal quality over prompting complexity. Parallel data, carefully filtered synthetic data, and well-structured preference signals consistently yield larger gains than complex reasoning or chain-of-thought style prompting.
    \item Preference-based training objectives should be treated as a central component of LLM based MT rather than as a post hoc refinement step, as they provide a more faithful mechanism for aligning translation outputs with human judgments.
    \item Fine-tuning strategies must be designed to preserve emergent capabilities of LLMs. Uncontrolled supervised fine-tuning risks reducing instruction-following ability, stylistic control, and contextual generalization, effectively collapsing models into brittle encoder–decoder behavior.
    \item Evaluation protocols should be diversified and human-centered. No single automatic metric captures all dimensions of translation quality, making multi-metric reporting necessary. In addition to numerical scores, evaluation methods should emphasize interpretability through structured error analysis and natural language explanations of judgments, enabling practitioners to understand why a translation is rated as good or poor. 
    \item Selective human evaluation remains essential, particularly for document-level and literary translation, where discourse coherence, stylistic adequacy, and creative fidelity are not reliably captured by automatic metrics alone.
    \item Progress in document-level MT requires explicit modeling of context. Long context windows alone are insufficient, and future systems should emphasize context selection, state tracking, and consistency control rather than treating documents as longer sentences.
\end{itemize}

Overall, LLMs do not replace traditional MT systems, but redefine the design space of MT. The most effective approaches combine the flexibility of LLMs with explicit supervision, preference-based alignment, and structured control mechanisms. As MT research increasingly intersects with instruction following, alignment, and evaluation, future progress will depend less on scale alone and more on how linguistic signal, optimization objectives, and human preferences are integrated into training and inference.

\bibliography{anthology,papers}% common bib file

\end{document}